\pdfoutput=1
\documentclass[11pt]{article}

\usepackage[]{ACL2023} 
\usepackage{times}
\usepackage{latexsym}
\usepackage[T1]{fontenc}
\usepackage[utf8]{inputenc}
\usepackage{microtype}
\usepackage{inconsolata}
\usepackage{booktabs}
\usepackage{graphicx}
\usepackage{multirow}
\usepackage{listings}

\title{Beyond Positive Scaling: \\ How Negation Impacts Scaling Trends of Language Models}

\author{
Yuhui Zhang\thanks{~~Equal contributions. $^\dag$Correspondence to: Yuhui Zhang <yuhuiz@stanford.edu>.}\:\:$^\dag$
\quad Michihiro Yasunaga$^{*}$
\quad Zhengping Zhou$^{*}$
\quad Jeff Z. HaoChen$^{*}$ \\ 
\textbf{
James Zou
\quad Percy Liang
\quad Serena Yeung
} \\
Department of Computer Science \\
Stanford University
}

\begin{document}
\maketitle
\begin{abstract}
Language models have been shown to exhibit positive scaling, where performance improves as models are scaled up in terms of size, compute, or data. In this work, we introduce NeQA, a dataset consisting of questions with negation in which language models do not exhibit straightforward positive scaling. We show that this task can exhibit inverse scaling, U-shaped scaling, or positive scaling, and the three scaling trends shift in this order as we use more powerful prompting methods or model families. We hypothesize that solving NeQA depends on two subtasks: question answering (task 1) and negation understanding (task 2). We find that task 1 has linear scaling, while task 2 has sigmoid-shaped scaling with an emergent transition point, and composing these two scaling trends yields the final scaling trend of NeQA. Our work reveals and provides a way to analyze the complex scaling trends of language models.
\end{abstract}

\section{Introduction}
\label{sec:introduction}

\begin{figure}[!t]
    \vspace{-0.8em}
    \centering
    \includegraphics[width=0.95\linewidth]{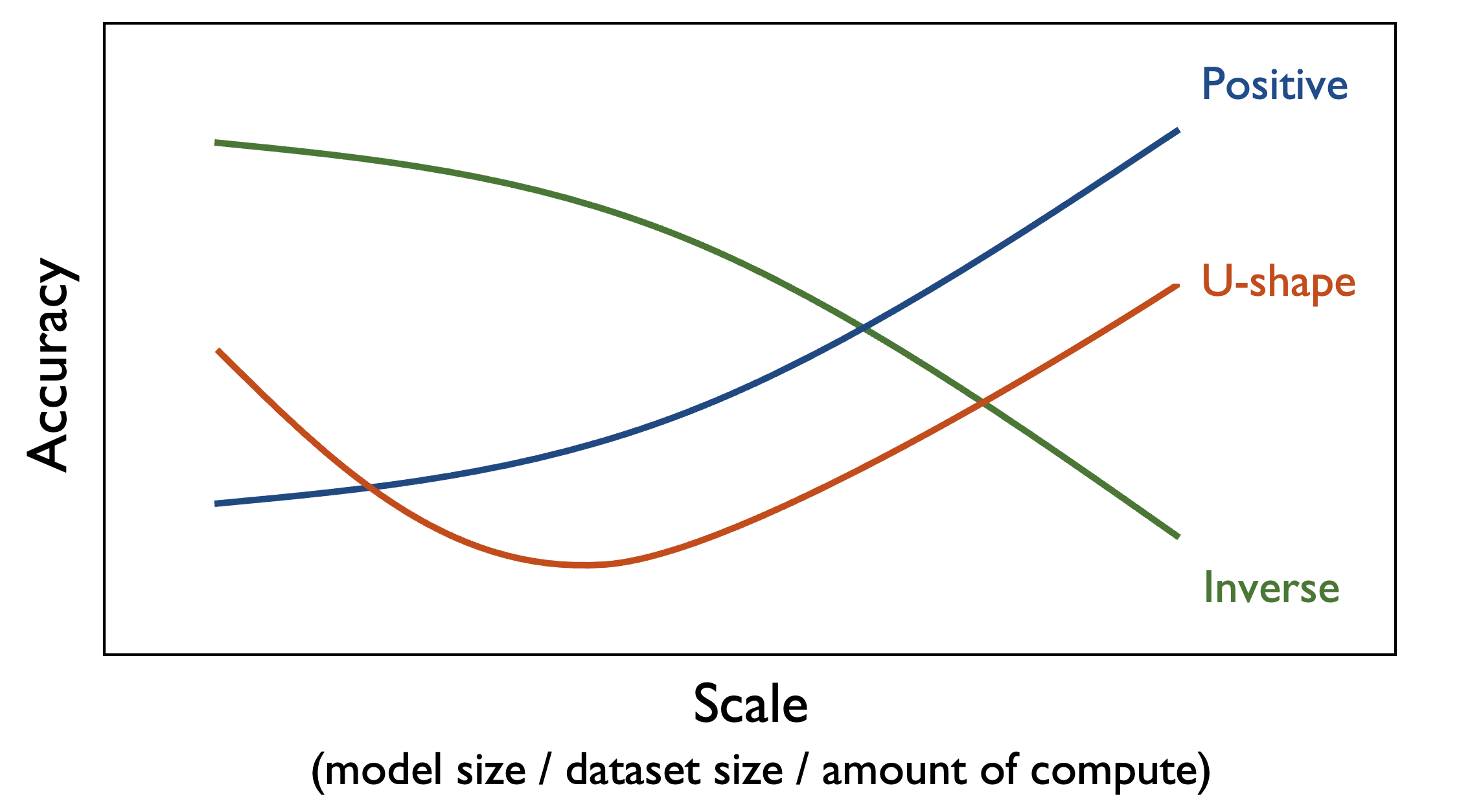}
    \vspace{-0.5em}
    \caption{Illustration of three types of scaling trends.}
    \vspace{-0.8em}
    \label{fig:scaling_demo}
\end{figure}

Language models have been shown to exhibit \textit{positive scaling}, where task performance improves as models are scaled up in terms of size, compute, or data, like the blue curve in Figure \ref{fig:scaling_demo}  \citep{kaplan2020scaling,brown2020language,rae2021scaling,chowdhery2022palm,srivastava2022beyond,liang2022holistic}. However, there are exceptions. Recent works show that some tasks can exhibit \emph{inverse scaling} \cite{mckenzie2022inverse}, where the performance degrades as models are scaled up (green curve), or \emph{U-shaped scaling} \cite{wei2022inverse}, where the performance degrades first but then improves as models are scaled up (red curve). Analyzing tasks that exhibit different scaling trends, such as inverse and U-shaped scaling, is therefore useful for better understanding the behaviors of language models, identifying their limitations, and guiding future development.

In this work, we introduce NeQA, a new task of answering multiple-choice questions containing negation words, constructed by transforming questions from OBQA~\citep{mihaylov-etal-2018-suit} and NegatedLAMA~\citep{kassner-schutze-2020-negated}. We conduct experiments on this task using 4 language model families and 3 prompting methods, and show that large language models do not follow straightforward positive scaling on this task. Specifically, as we use more powerful prompting methods or model families, NeQA exhibits a gradation from inverse scaling to U-shape to positive scaling. This result provides a unified view of when the three types of scaling trends (inverse, U-shaped, and positive scaling) occur for language models. Our result indicates that the development of large language models' capability to process negation may be a complex and nuanced problem.

\begin{figure*}[htbp]
    \vspace{-5mm}
    \begin{minipage}[htbp]{0.5\linewidth}
        \centering
        \includegraphics[width=0.9\linewidth]{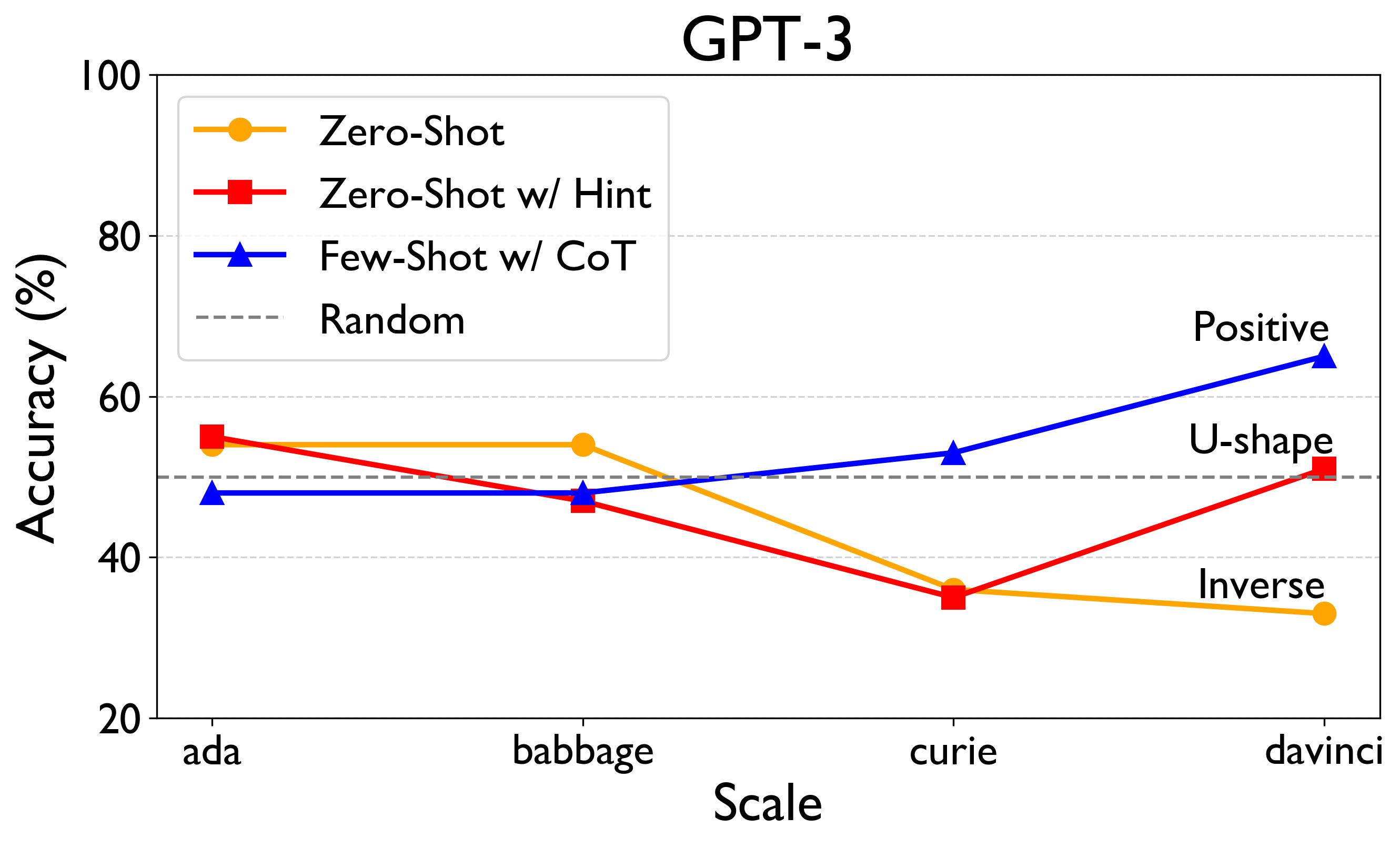}
    \end{minipage}
    \begin{minipage}[htbp]{0.5\linewidth}
        \centering
        \includegraphics[width=0.9\linewidth]{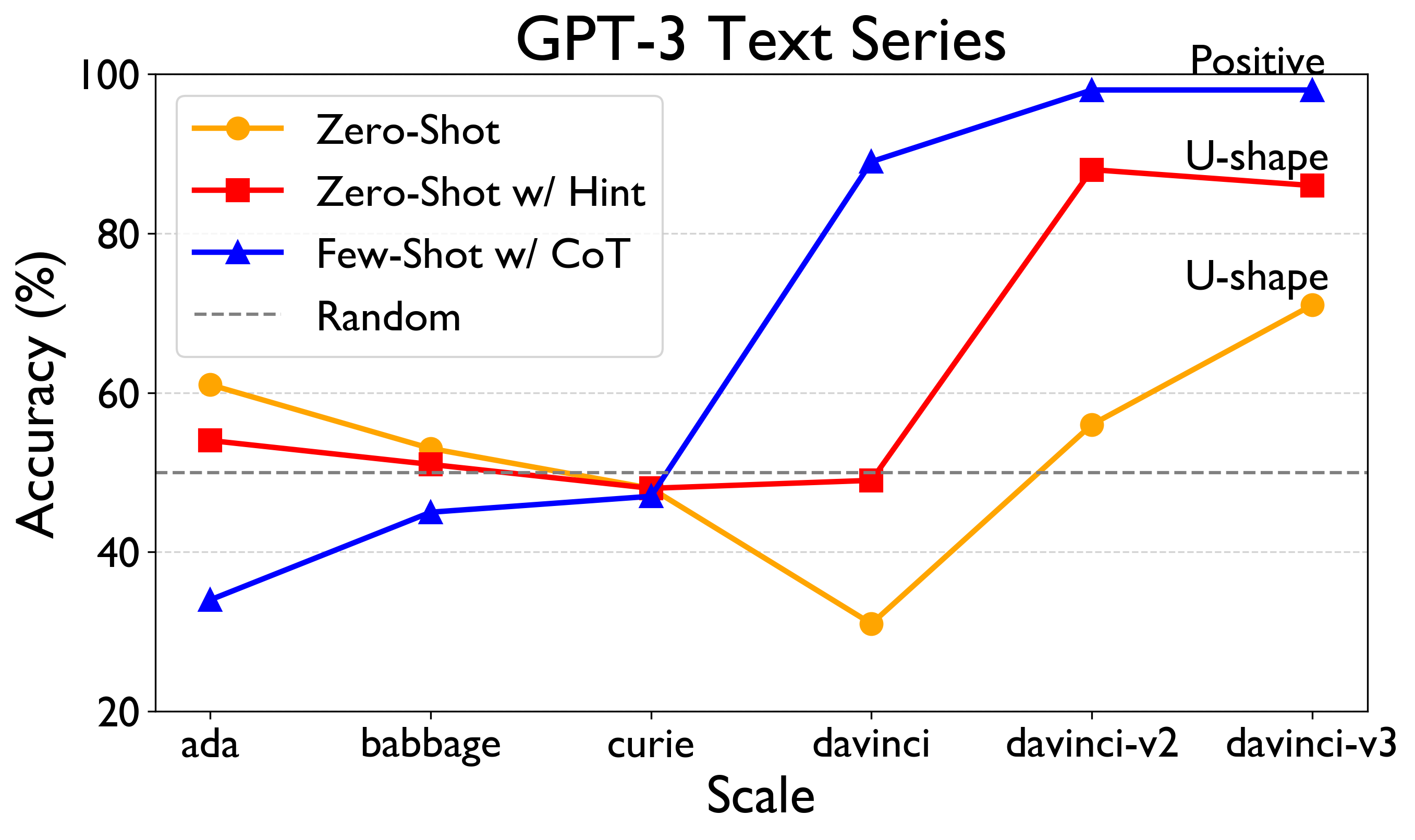}
    \end{minipage}
    \begin{minipage}[htbp]{0.5\linewidth}
        \centering
        \includegraphics[width=0.9\linewidth]{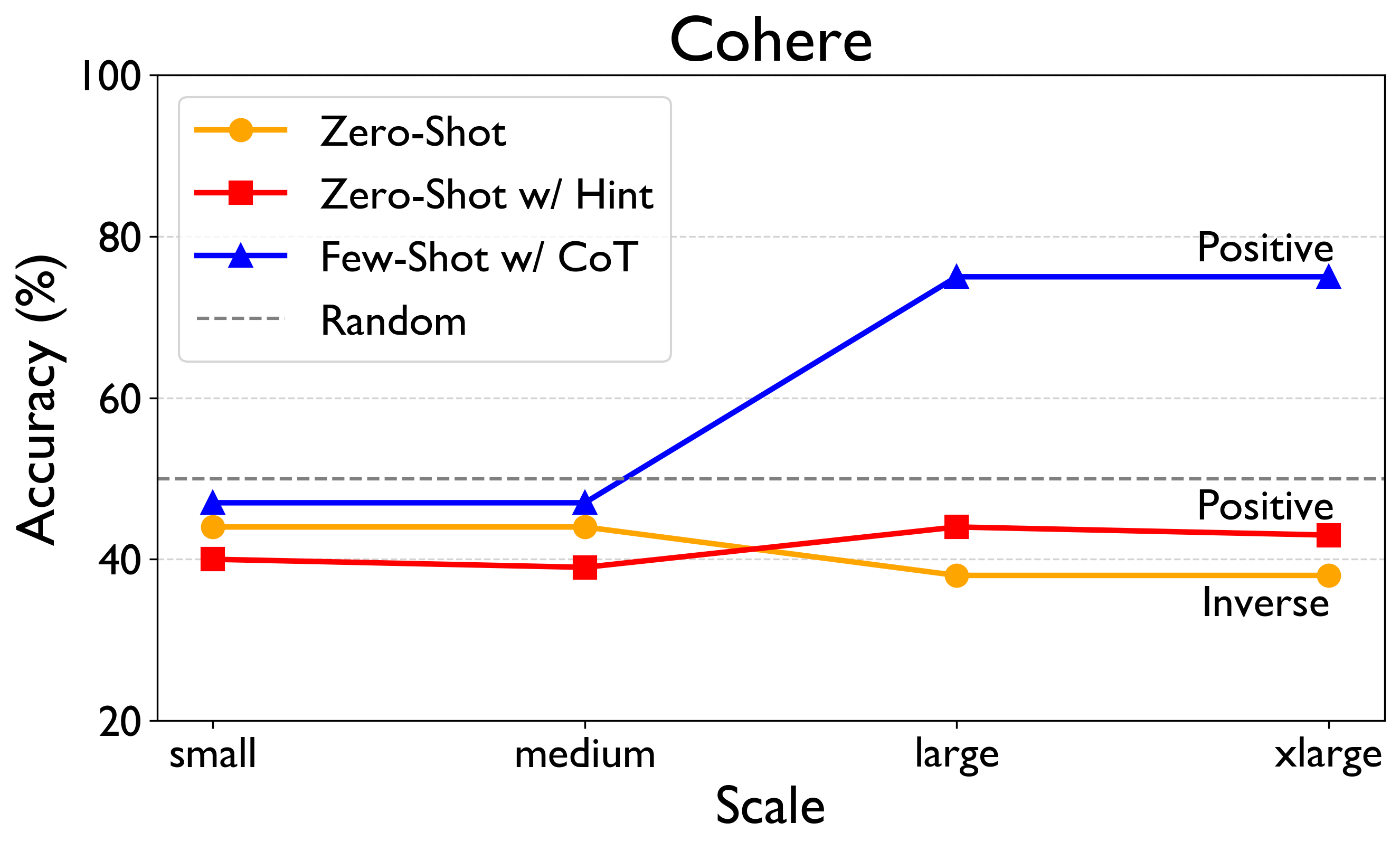}
    \end{minipage}
    \begin{minipage}[htbp]{0.5\linewidth}
        \centering
        \includegraphics[width=0.9\linewidth]{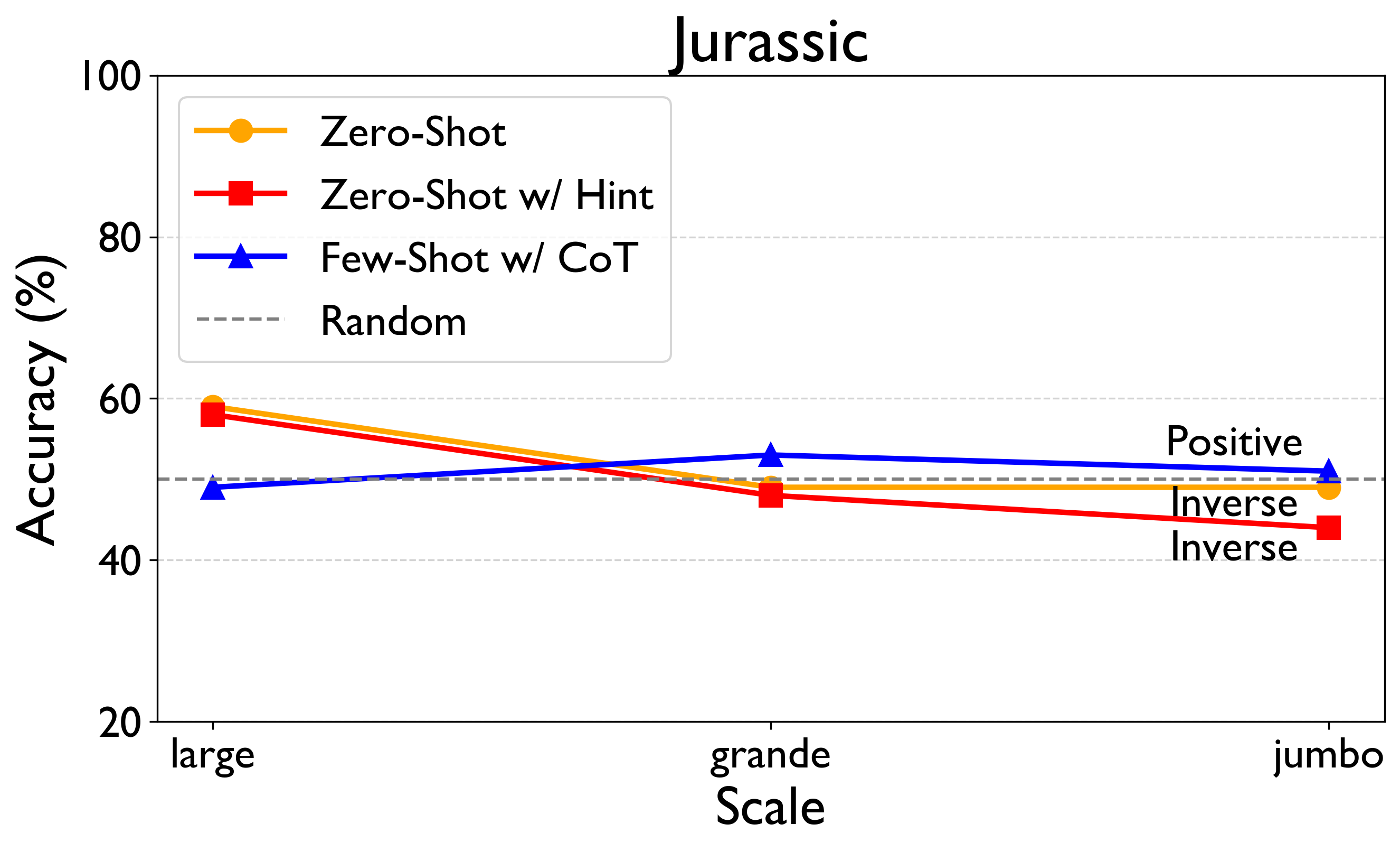}
    \end{minipage}
    \caption{Scaling trends of various language models on the NeQA dataset. As we use more powerful prompting methods or model families, we observe a gradation from inverse scaling to U-shape to positive scaling. More details in \S \ref{sec:exp}. } 
    \label{fig:scaling}
    \vspace{-1.5mm}
\end{figure*}

To further understand this nuanced scaling trend of the NeQA task, we decompose the task into two subtasks: question answering (task 1) and negation understanding (task 2). Our empirical results show that task 1 has linearly positive scaling, while task 2 has sigmoid-shaped scaling with an emergent transition point, where the transition point is influenced by the prompt method and model family. Combining these two scaling trends yields the final scaling trend observed in NeQA. The task decomposition provides a new way to think of the scaling on a task in terms of a combination of its component skills.

In summary, our contributions are (1) the NeQA dataset that contains diverse distributions of texts about negation; (2) an evaluation of different large language models on the NeQA dataset, which exhibits different scaling trends; (3) a task decomposition analysis explaining the above scaling trends. 

\section{Dataset: NeQA}
\label{sec:dataset}

We develop NeQA, a question answering dataset designed to evaluate the ability of models to process negation in natural language. Each example of the dataset consists of a negated question and two answer choices, one correct and one incorrect. An example of NeQA looks like: (question ``Child does not want?'', correct choice ``marriage'', incorrect choice ``love''). To construct this, we leveraged NegatedLAMA~\citep{kassner-schutze-2020-negated} and OBQA~\citep{mihaylov-etal-2018-suit}. 

\begin{figure*}
\begin{centering}
    \vspace{-5mm}
    \centering
    \includegraphics[width=\linewidth]{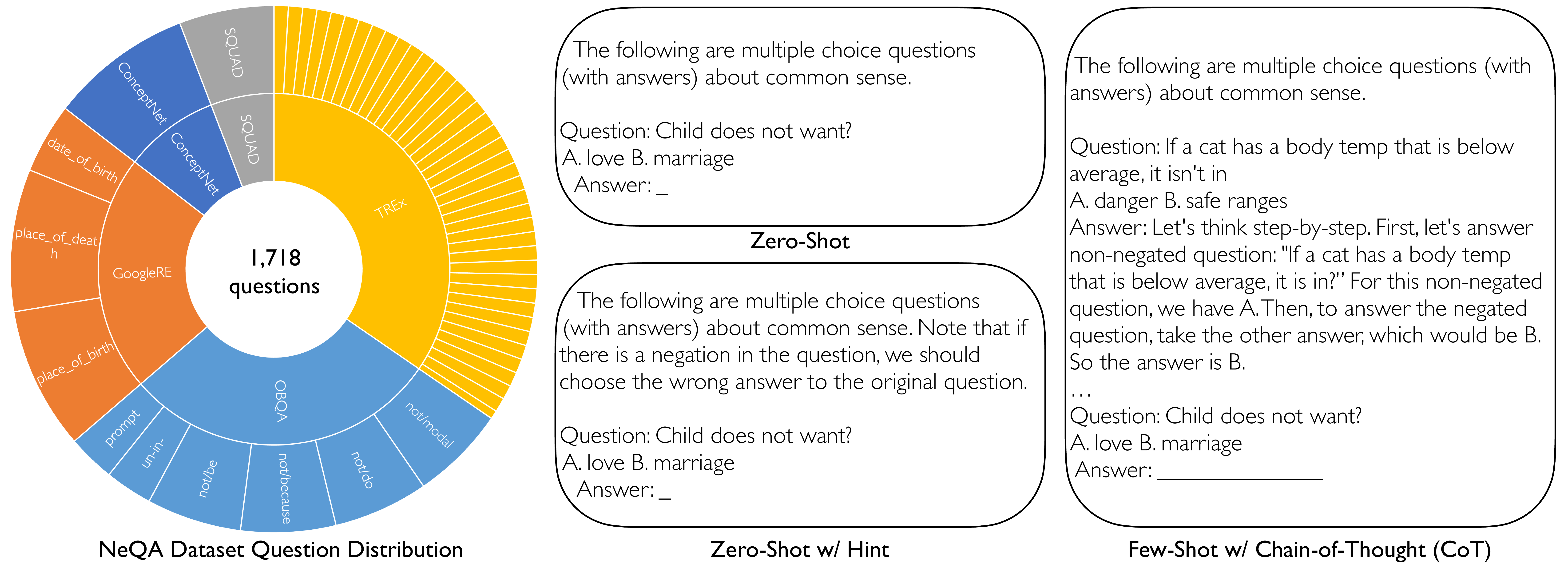}
    \caption{(Left) Statistics of NeQA data sources. (Right) Three prompting methods that yield shifts in scaling trends.}
    \label{fig:data_and_prompt}
\end{centering}
\vspace{-1.5mm}
\end{figure*}

The NegatedLAMA dataset includes negated questions from four subsets: ConceptNet, GoogleRE, SQuAD, and TREx. Each subset comprises multiple files that represent different question distributions, such as questions about different entity relations. Each question is associated with a negated question, an answer, and a misprimed question (i.e., a wrong answer followed by the question). For instance, when ``Child wants?'' is the original question, ``Child does not want?'' can be its associated negated question, ``love'' can be the answer, and ``Marriage? Child wants?'' can be its misprimed question. We turn it into a multiple choice question by setting the negated question as the question, and setting the wrong answer in the misprimed question in conjunction with the correct answer as the two choices. For instance, in the above example, we get ``Q: Child does not want? A. love B. marriage'' (Appendix Table~\ref{tab:data_generation_example}). To ensure diversity and representativeness, we randomly selected at most 50 questions from each file.

To be able to analyze the impact of different negation types, we also created additional data by applying diverse rules to transform questions in OBQA~\citep{mihaylov-etal-2018-suit} into negated ones. We defined six types of negation transformations: action verb negation (e.g., ``cause'' $\rightarrow$ ``does not/doesn't cause''), linking verb negation (e.g., ``is'' $\rightarrow$ ``is not/isn't''), modal verb negation (e.g., ``can'' $\rightarrow$ ``can not/can't''), conjunction negation (e.g., ``because'' $\rightarrow$ ``not because''), negation prefix (e.g., ``able'' $\rightarrow$ ``unable''), and negation prompt (e.g., add ``choose the wrong answer''). For each type, we collected 50 questions by applying a rule-based transformation, sampling an incorrect answer as the correct answer, and treating the correct answer as the incorrect answer. For example, ``Pushing on a pedal is an example of'' is an original question in OBQA with the correct answer ``force'' and one of the incorrect answers ``speed''. We apply the rule-based transformation to change the verb ``is'' to ``isn't'' and get ``Q: Pushing on a pedal isn't an example of? A. speed. B. force'', where ``A'' is the answer (Appendix Table~\ref{tab:data_generation_example}).

We employ post-processing techniques such as redistributing labels evenly between ``A'' and ``B'' and balancing the use of negation words such as ``not'' and ``n't''. The validity of each question is ensured through manual examination and editing. Our dataset comprises a total of 1718 questions sourced from ConceptNet (150 questions), GoogleRE (374 questions), SQuAD (100 questions), TREx (594 questions), and OBQA (500 questions), providing a diverse range of negation types, text distributions, and prompts. We believe that this dataset serves as a valuable benchmark for assessing the ability of language models to process negation. Data distributions are shown in Figure~\ref{fig:data_and_prompt}. 

Out of the 1718 questions, we define a set of 944 questions from ConceptNet, TREx, and a subset of OBQA that exhibit clear positive scaling on the corresponding original (non-negated) questions. For our experiments (\S \ref{sec:result}), we randomly select 100 questions from this positive set in order to make the scaling more obvious during our analysis.

\section{Results}
\label{sec:result}

\begin{figure*}
\begin{centering}
    \vspace{-4mm}
    \centering
    \includegraphics[width=\linewidth]{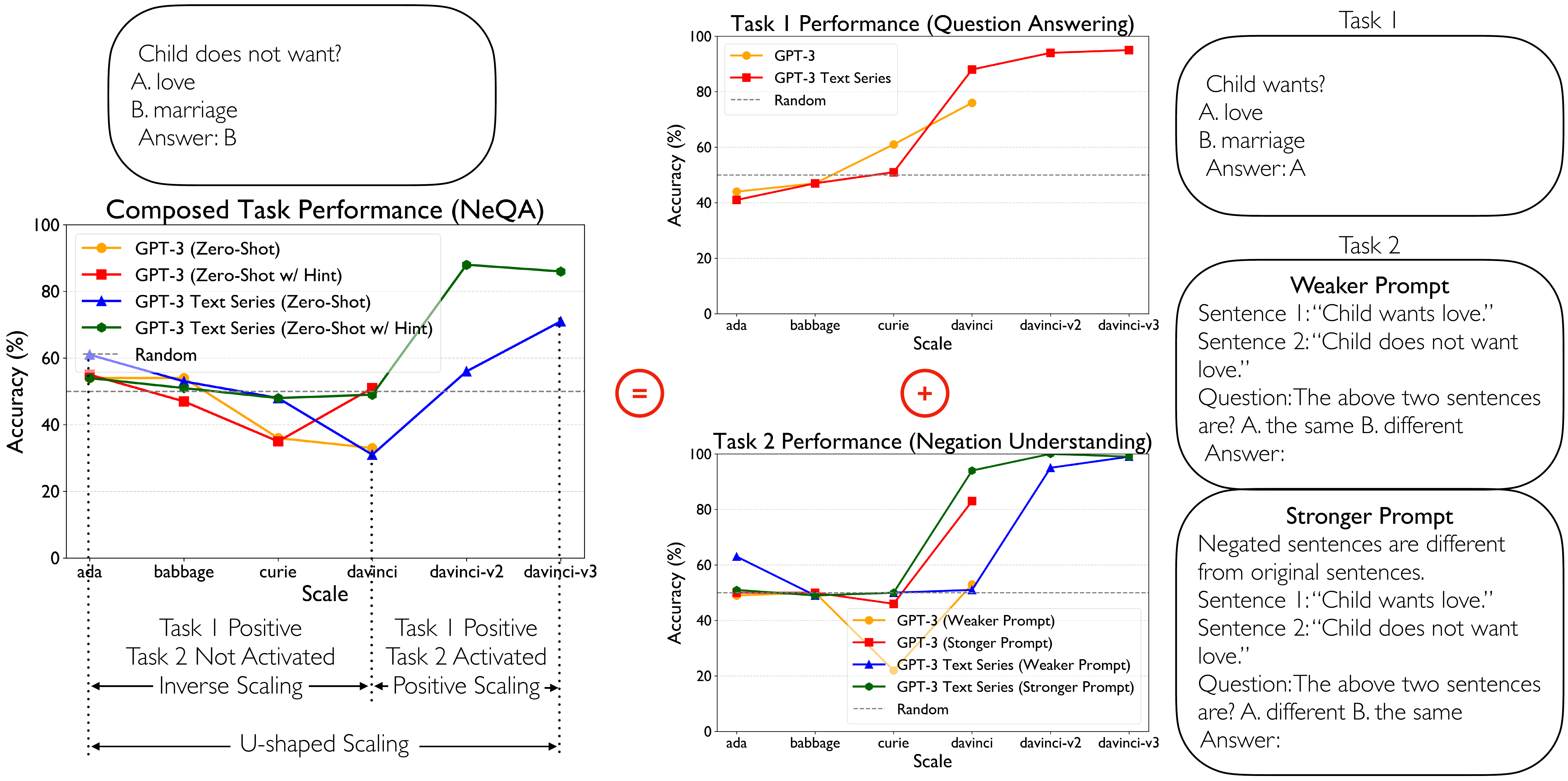}
    \caption{
    Task decomposition analysis of the NeQA task. 
    NeQA task can be decomposed into two subtasks: question answering (task 1) and negation understanding (task 2). Our empirical results show that task 1 has linear scaling, while task 2 has sigmoid-shaped scaling with an emergent transition point, where the transition point is influenced by the prompt method and model family. Combining these two scaling trends yields the final scaling trend observed in NeQA.
    }
    \label{fig:task}
\end{centering}
\vspace{-1.5mm}
\end{figure*}

\subsection{Evaluation Setup: Models and Prompts}

We evaluated four different language model families on NeQA: GPT-3~\cite{brown2020language}, GPT-3 Text Series~\cite{ouyang2022training}, Cohere~\cite{cohere}, and Jurassic~\cite{jurassic} (model details in \S \ref{sec:models}). We employed three different prompting methods: zero-shot, zero-shot with hint~\cite{kojima2022large}, and few-shot with chain-of-thought (CoT)~\cite{wei2022chain}, as illustrated in Figure~\ref{fig:data_and_prompt}. 

For zero-shot and zero-shot with hint evaluation, we follow the evaluation protocol of the MMLU paper~\cite{hendrycks2020measuring}. We generate a prompt composed of a question and multiple choice options, where the options are labeled ``A'' and ``B''. For example, a prompt may be ``Question: Child does not want? A. love B. marriage Answer:''. We then generate one token from the language model and rank the probability of the model selecting option ``A'' or ``B''. For few-shot with CoT, we follow the evaluation protocol of CoT paper~\citep{wei2022chain} by generating sentences until reaching the end and parsing the answer using regular expressions. As our metric, we report the accuracy of the model predictions, where the chance accuracy is 50\% as NeQA is a balanced two-choice dataset.

\subsection{Scaling Trends}

Our evaluation reveals that the scaling trends of language models on the NeQA task vary depending on the prompting method and model family used (Figure~\ref{fig:scaling}). We found that the scaling trends of all language model families can be altered by different prompts. For example, zero-shot prompting resulted in inverse scaling in 3 out of 4 model families, whereas few-shot CoT prompting consistently resulted in positive scaling. As the prompt becomes stronger (i.e., more information, like rationales and demonstrations, is provided for the language model), we observed a transition from inverse scaling, to U-shaped scaling, to positive scaling. For instance, GPT-3 exhibited inverse scaling, U-shaped scaling, and positive scaling, respectively, with these prompting methods. Additionally, we discovered that switching to a stronger model family can alter the scaling shape. For example, transitioning from GPT-3 to GPT-3 Text Series, which was further trained to align with human values on multiple tasks compared to GPT-3, resulted in a shift from inverse scaling to U-shaped scaling when the same prompting (e.g., zero-shot) is used. 

In conclusion, stronger prompts or model families lead to a transition from inverse scaling, to U-shaped scaling, to positive scaling on the NeQA task. We may also make the following interpretation: the overarching scaling trend of language models for NeQA is U-shaped, and if the model is weak (i.e., weaker prompt or model family), the left part of ``U''/inverse slope is observed; if the model is strong, the right part of ``U''/positive slope is observed.

\subsection{Task Decomposition Analysis}

We conducted further empirical analysis on the reasons why the scaling trends can be inverse, U-shaped, or positive and can transition with different prompts or model families. We decomposed the NeQA task into two subtasks: task 1 is to answer the original non-negated questions, and task 2 is to ``understand negation''. In Figure~\ref{fig:task}, we show the scaling of task 1 and task 2 performance with GPT-3 and GPT-3 Text Series families. The task 1 performance is measured by the accuracy of answering original non-negated questions, and the task 2 performance is measured by the accuracy of differentiating original questions from negated questions. The task examples are shown in Figure \ref{fig:task} right. Both tasks are evaluated in a zero-shot way. 

Our experiments showed that task 1 scales mostly linearly in a positive direction, whereas task 2 scales like a sigmoid shape with an emergent transition point, analogous to the Grokking curve \cite{power2022grokking}. Before this transition point, models do not ``understand negation'', and achieve low accuracy in differentiating original questions from negated questions, which results in outputting the same answer to both the original and negated questions. It is worth noting that the labels for the composed task NeQA are essentially the inverse of the non-negated QA labels for task 1. Therefore, the positive scaling in task 1 results in inverse scaling for the composed task NeQA, because the predictions remain unchanged while the ground-truth labels are inverted. After the transition point, models start to ``understand negation'' and predict opposite answers to the original questions, resulting in positive scaling. When the transition point never happens within the sizes available in the model family, the overall scaling looks inverse; when the transition point happens before the smallest model, the overall scaling looks positive. When the transition point is in the middle, the overall scaling looks U-shaped. We provide further explanations of the composed performance curve in \S \ref{sec:decomposition} and \S \ref{sec:finetuning}.

Interestingly, we found that the transition point can be moved earlier with stronger prompting methods or model families. For example, both GPT-3 and GPT-3 Text Series show that the transition point happens much earlier when using the stronger prompt compared to the weaker prompt (see Figure~\ref{fig:task}). Furthermore, GPT-3 Text Series has an earlier transition point than the GPT-3 models. This can explain why using stronger prompts or stronger model families results in a transition from inverse scaling, to U-shaped, to positive scaling.

By decomposing a task and studying the scaling trends of the individual subtasks, our analysis offers a new way to understand the complexity of language model scaling trends. This analysis could be applied to various tasks beyond NeQA, especially tasks that consist of multiple subtasks, each of which may be of different levels of difficulty. This analysis can provide a deeper understanding of the strengths and weaknesses of different language models and offer useful insights into the development of better models and training/prompting methods.

\section{Related Works}
\label{sec:related_works}

\paragraph{Scaling trends.}

Recent years have seen significant scaling of language models, such as scaling from GPT-1 to GPT-3, which has led to tremendous improvements in their performance and capabilities in natural language processing~\cite{radford2018improving,radford2019language,brown2020language}. Researchers have begun to investigate the scaling trends of language models to capture the relationship between model performance and model scale, including the parameter count and amount of training data/compute used~\citep{kaplan2020scaling}. While most scaling papers show positive scaling trends where larger models perform better on various tasks~\cite{brown2020language,rae2021scaling,chowdhery2022palm,srivastava2022beyond,liang2022holistic}, it is important to also investigate tasks that exhibit other trends such as inverse scaling, which can shed light to limitations in current language model development and guide future improvements. For instance, TruthfulQA~\cite{lin-etal-2022-truthfulqa} was one of the earliest tasks that exhibit inverse scaling, where they find larger language models are prone to hallucination and generate more untrue answers. A recent competition, the Inverse Scaling prize~\cite{mckenzie2022inverse}, called for tasks that cause inverse scaling. In the first round, four tasks, including NeQA, redefine math, quote repetition, and hindsight neglect, showed inverse scaling. \citet{wei2022inverse} then found that some of these tasks show U-shaped scaling after further scaling up language models. In this work, we unify the above findings and provide a holistic picture of scaling trends, including the transition from inverse to U-shaped to positive scaling across model families and prompting methods, and empirical explanations behind these scaling trends.

\paragraph{Negation understanding.}

Negation is a fundamental aspect of natural language understanding \cite{ackrill1975categories,blanco-moldovan-2011-semantic}. Existing works have found that NLP models can struggle in processing negation in text \cite{jimenez-zafra-etal-2020-corpora}. For example, these works investigate models' abilities to process negation through natural language inference tasks \cite{cooper1996using,dagan2006pascal,hossain-etal-2020-analysis,geiger-etal-2020-neural}, machine translation \cite{fancellu-webber-2015-translating-negation,hossain-etal-2022-analysis}, language model prompting \cite{kassner-schutze-2020-negated,ettinger-2020-bert,jang2022can}, contrastive reading comprehension \cite{ravichander2022condaqa}, and probing model activations \cite{burns2022discovering}. In response, existing works have also studied methods to improve the abilities of NLP models to process negation, such as leveraging datasets about negation \cite{kim-etal-2019-probing,jiang-etal-2021-im}, auxiliary training objectives/tasks \cite{khandelwal2019negbert,moore2020multi,hosseini2021understanding,truong2022improving}, and neuro-symbolic reasoning modules \cite{yasunaga2021qa, yasunaga2022dragon}. While these existing works typically study a fixed size or type of models, our work provides the first studies into the effect of negation on the {\textit{scaling trends}} of language models. We find that negation can exhibit nuanced scaling trends, e.g., U-shaped scaling with increased model size and improved model families and prompting methods. This finding offers a more comprehensive insight into how to improve the abilities of language models to understand negation, e.g., the model size, training algorithm, and prompting method all matter.

\section{Conclusion}
\label{sec:conclusion}

We introduced NeQA, a new question answering dataset that yields different scaling trends of language models than traditional positive scaling. We then proposed task decomposition analysis, a general idea to decompose the task to better understand the complex scaling trends and their transitions. We hope that these insights can facilitate the understanding and development of language models.

\section*{Limitations}

This work introduced NeQA, a question answering dataset for evaluating the ability of large language models to process negation. While our NeQA attempted to cover diverse types of negation (e.g., different negation phrases and positions) and multiple data sources (e.g., OBQA, LAMA), it is possible that the dataset construction misses some types of negation or domains of text. Our future work will extend the dataset to cover more comprehensive types of negation and domains of text, beyond OBQA and LAMA. Additionally, NeQA is an English dataset, and it would be interesting to extend it to non-English languages and conduct a more comprehensive evaluation of language models, including multilingual ones. 

Another potential limitation is sensitivity in language model prompting. Language model performance is known to be influenced by the specific prompt used to query the model (e.g., a rephrased prompt may lead to different model outputs), and prompt engineering---finding the ``right'' prompt---may be needed to obtain reasonable outputs from the language models \cite{jiang-etal-2020-know,ruis2022large,wang2022self}. As our language model evaluation protocol uses prompting (\S \ref{sec:result}), the evaluation results may inherit such prompt sensitivity. It would be an interesting future work to incorporate techniques to mitigate prompt sensitivity in language model evaluation (e.g., \citealt{burns2022discovering}).

\section*{Ethics Statement}

Our work offers benchmarks and insights to help develop language models that understand negation. Developing language models that understand negation is crucial to the society in many ways.

First, as language models are being used in various real-world applications, including fields like finance, healthcare, and law, it is important to ensure that they understand negation and make correct predictions. If they do not understand negation, they may output the opposite of what we actually want and may make harmful decisions for humans.

Negation is also a fundamental aspect of natural language understanding, and a language model that does not understand negation correctly may not be able to truly process natural language. This can undermine trust and confidence in the outputs of the model, ultimately undermining its utility.

Understanding negation correctly is therefore crucial for the development of reliable language models. We hope that our benchmark and evaluation results provide insights into the behavior of current language models and inspire the future development of language models that understand negation.

\section*{Reproducibility Statement}

We provide our datasets and implementations at \url{https://github.com/yuhui-zh15/NeQA}. The implementations will enable researchers to reproduce datasets and results described here, as well as apply our negation transformations to other datasets and run their own analyses.

\section*{Acknowledgments}

We greatly thank members of Stanford NLP, P-Lambda, and MARVL groups for providing valuable feedback. M.Y. is supported by Microsoft Research PhD Fellowship. J.Z. is supported by a Sloan Fellowship and a Chan-Zuckerberg Investigator Award. P.L. is supported by a PECASE Award. S.Y. is supported by a Chan-Zuckerberg Investigator Award.

\newpage
\bibliography{anthology,custom}
\bibliographystyle{acl_natbib}

\appendix
\newpage

\section{Task Decomposition Simulation: Composing Subtask Scaling Trends Yields U-shape Scaling}
\label{sec:decomposition}

\begin{figure}[htbp]
    \centering
    \includegraphics[width=\linewidth]{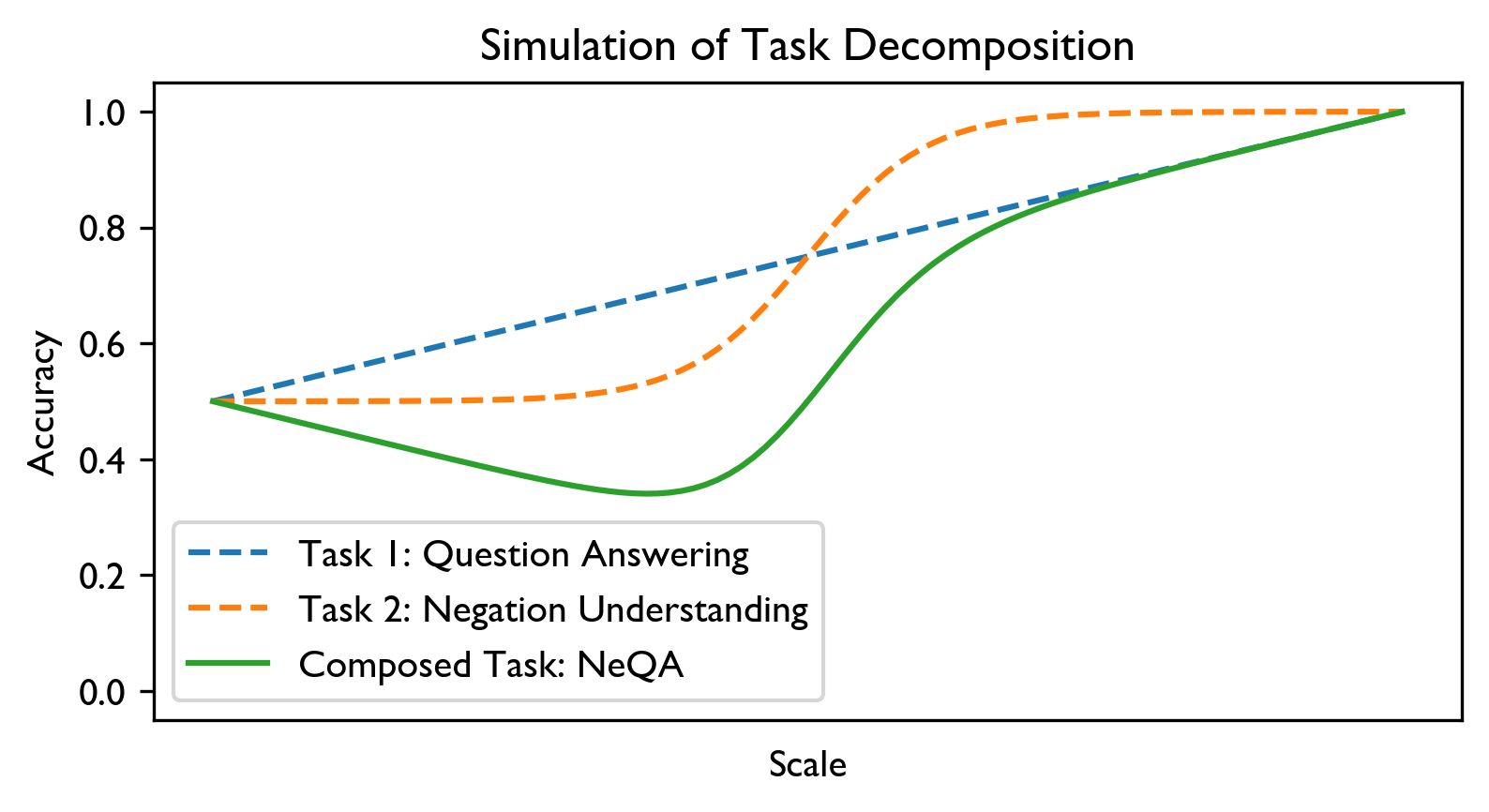}
    \caption{Task decomposition simulation illustrates the U-shape of composed task when given individual tasks.}
    \label{fig:task_synthesis}
\end{figure}

\begin{figure*}[htbp]
    \begin{minipage}[htbp]{0.5\linewidth}
        \centering
        \includegraphics[width=\linewidth]{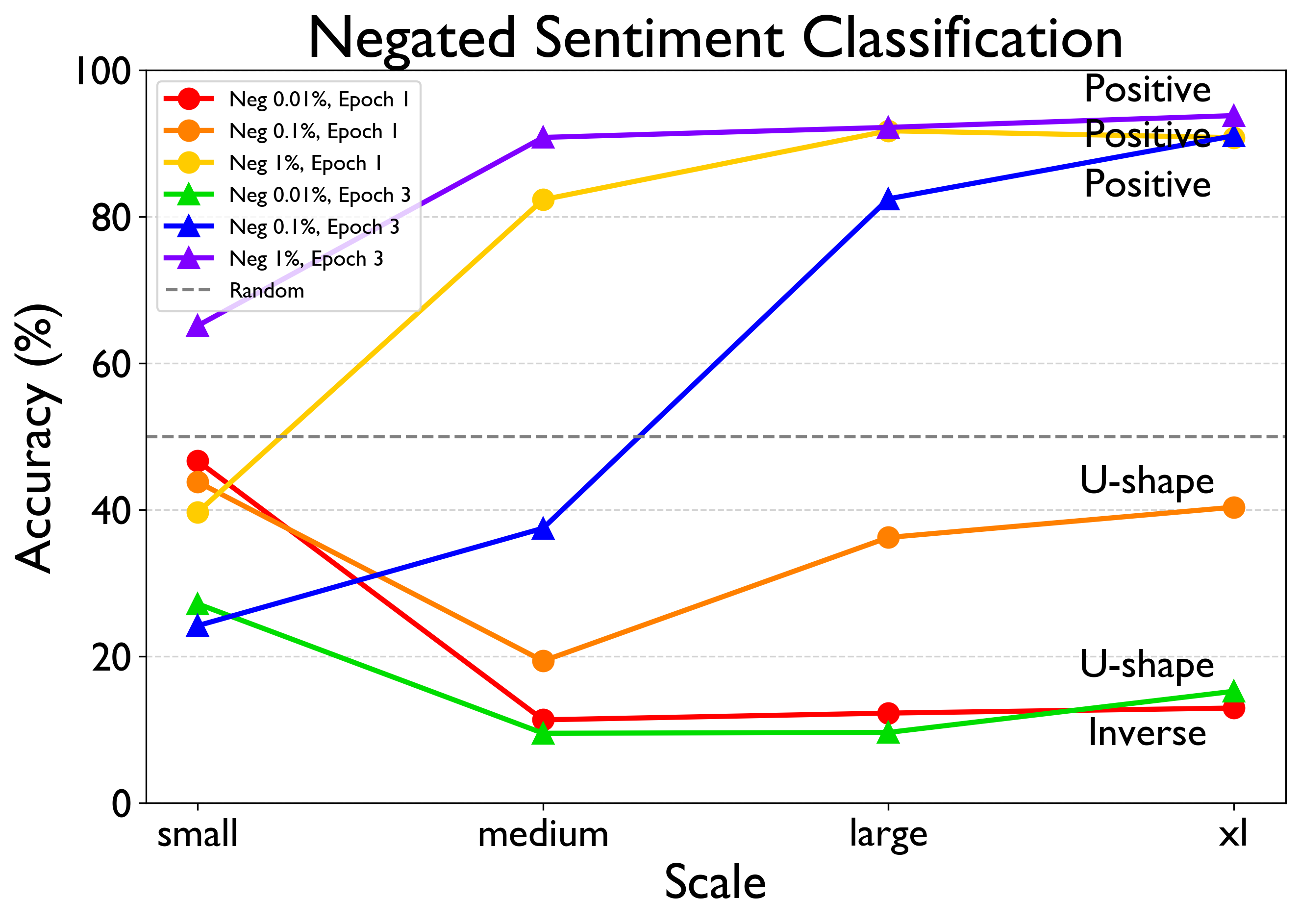}
    \end{minipage}
    \begin{minipage}[htbp]{0.5\linewidth}
        \centering
        \includegraphics[width=\linewidth]{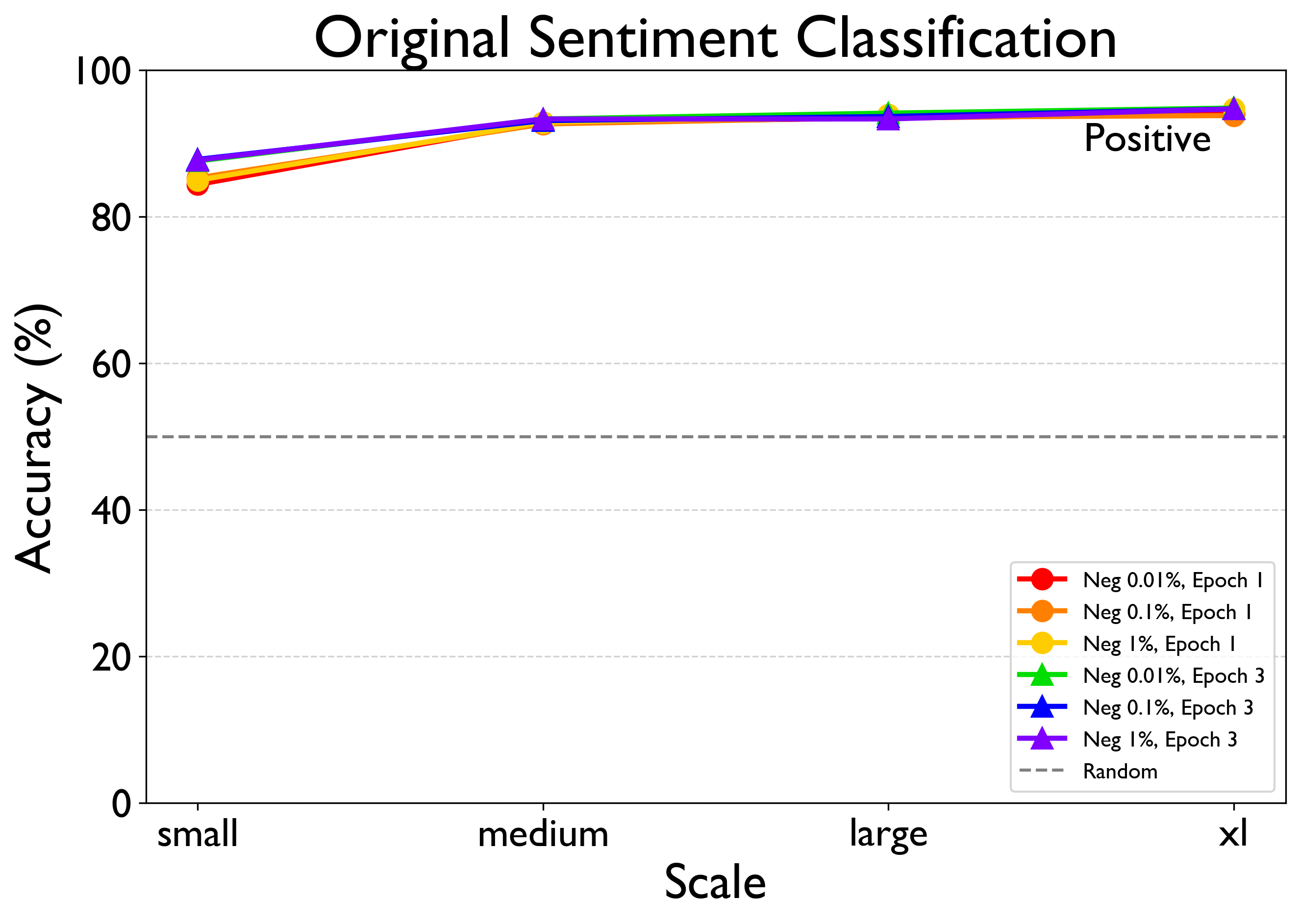}
    \end{minipage}
    \caption{Fine-tuning simulation reveals dataset attributes and training computes also impact scaling trends. We fine-tune different-sized GPT-2 models on a transformed SST-2 sentiment classification dataset with the casual language modeling objective. When the negation ratio in the dataset or fine-tuning epoch increases, we observe a shift from inverse scaling to U-shape to positive scaling on negated sentiment classification task. The original sentiment classification task always shows positive scaling. }
    \label{fig:finetuning}
\end{figure*}

In this section, we present a simple simulation to demonstrate how the U-shape scaling trends of a composed task can be obtained through the scaling trends of each decomposed task. Let's assume that the accuracy of Task 1 (Question Answering) is represented by $t_1(x)$ and has a linear shape with an initial performance of 0.5 (random performance) and a final performance of 1.0 (perfect performance). Similarly, the accuracy of Task 2 (Negation Understanding) is represented by $t_2(x)$ and has a sigmoid-like shape with an initial performance of 0.5 (random performance) and a final performance of 1.0 (perfect performance), where $x$ represents the scale (a combination of model size, data size, training computation, and prompting method). We define the score of negation understanding as $s_2(x) = (t_2(x) - 0.5) / 0.5$, which represents the probability that the model will treat a negated sentence differently from the original sentence. For the composed task, NeQA, it will have an accuracy of $t(x) = t_1(x)s_2(x) + (1 - t_1(x)) (1-s_2(x))$ given scale $x$.

Figure~\ref{fig:task_synthesis} shows the plots of these three curves, $t_1(x)$, $t_2(x)$, and $t(x)$. The simulated performance curve of NeQA, $t(x)$, indeed exhibits a U-shape.

\paragraph{Discussion of Task Decomposition Validity and Generalizability to Other Tasks.}

We first clarify that task decomposition analysis is not intended to derive scaling laws (i.e., predict the exact performance of language model scaling). Instead, our analysis aims to explain scaling trends (inverse, U-shape, positive). For example, translation performance may not be simple addition of generation performance and word translation performance but should be positively correlated. Furthermore, while this exact decomposition structure might not hold in more complex tasks, our proposed decomposition analysis is a pioneering attempt to explain scaling trends on a task other than vanilla language modeling. Investigating the applicability of decomposition to other tasks is an essential future direction, and we hope our work will inspire others to push these boundaries. Lastly, we believe that our work's focus on negation is already a well-scoped and significant research contribution, as negation is one of the most common linguistic phenomena. To study negation, we collected the NeQA dataset, which exhibits inverse/U/positive scaling. To explain this, we propose this decomposition intuition, which works well because answering negated questions requires first answering the original questions and then flipping the answers.

\section{Fine-tuning Simulation: Training Data Attributes and Training Computes Also Impact Scaling Trends}
\label{sec:finetuning}

In addition to the prompting methods and model families that we studied in \S \ref{sec:result}, we are also interested in studying other factors that may contribute to scaling trends, specifically those related to the \emph{training} process. However, most large language models are not publicly available and training/reproducing them from scratch would require excessive computational resources. In light of this, here we conduct experiments using synthetic data and small-size language models to simulate and analyze the language model learning process.

We adapt the SST-2 dataset~\cite{socher-etal-2013-recursive} for our simulation. For each sentence $s$ in SST-2, with probability $1-x$, we modify it to ``$s$. This does suggest it is good/bad (depending on the label)'', and with probability $x$, we change it to ``$s$. This does not suggest it is {good/bad}''. Then, we fine-tune different sizes of GPT-2~\cite{radford2019language} on this synthetic corpus with the standard causal language modeling objective. We vary the numbers of epochs $t$ and negation ratio $x$ to understand their effect to scaling trends.

To evaluate the fine-tuned language models, we use the language model to complete ``$s$. This does suggest it is \_'' for the original sentiment classification task (similar to task 1 in the main paper), and use the language model to complete ``$s$. This does not suggest it is \_'' for the negated sentiment classification task (similar to the composed task NeQA in the main paper). We report accuracy on the original sentiment classification and negated sentiment classification.

Our simulation demonstrates that the scaling trends on negated sentiment classification are influenced by the negation ratio $x$ and training epoch $t$ (Figure~\ref{fig:finetuning}). With the same number of training epochs $t=1$, increasing the negation ratio $x$ from 0.01\%, to 0.1\% and then to 1\% causes the scaling to shift from inverse scaling, to U-shape, then to positive scaling. Additionally, increasing the number of training epochs from 1 to 3 causes the scaling trend to shift from inverse scaling to U-shape when the negation ratio is $x=0.01\%$, and from U-shape to positive when the same negation ratio is $x=0.1\%$.

This simulation highlights that factors in the \emph{training} process, such as dataset attributes (e.g., negation ratio) and training compute, also have significant impacts on the scaling trends. Together with the \emph{inference} factors, such as prompting methods and model families discussed in the main paper, we provide a comprehensive understanding of the complexity of scaling trends and how different factors can influence them. 

The transition of the scaling trends can also be explained by task decomposition, where Task 1 (original sentiment classification) is always positively scaled, while Task 2 (negation understanding) is also positive but is shaped like a sigmoid, with the transition point controlled by the number of negation examples seen by the language model. The number of negations seen can be modified by using a larger negation ratio or more training epochs. The composition of these subtask scaling trends yields the final scaling curves. 

The reason why Task 1 has a more linear shape, while Task 2 has a more sigmoid-like shape, can be understood with the intuition of deep learning processes. Empirical risk optimization (ERM) optimizes for average performance, and since negated sentences are significantly underrepresented in comparison to non-negated sentences in the training data, they are ignored at the beginning of training~\citep{sagawa2019distributionally,sohoni2020no,liu2021just}. As a result, the performance for negated sentences lags behind the average. However, as the majority of the training examples are learned, ERM finally starts to optimize for the underrepresented groups, leading to improved performance for negated sentences. This intuition adds new insights into the emergence of language models~\cite{bommasani2021opportunities,weiemergent}, and we leave more rigorous analyses to future works.

\section{Experimental Details}
\label{sec:exp}

\subsection{Results}
\label{sec:performance}

The performance of various models on different tasks that generate Figure~\ref{fig:scaling} and Figure~\ref{fig:task} can be found in Table~\ref{tab:performance} and Table~\ref{tab:performance-2}.

\begin{table*}[htbp]
\centering
\small

\begin{tabular}{ccc|cccc}
\toprule
Model & Prompting & Shape & ada & babbage & curie & davinci \\
\midrule
\multirow{3}{*}{GPT-3} & Zero-shot & Inverse & 0.54 & 0.54 & 0.36 & 0.33 \\
& Few-shot & Inverse & 0.51 & 0.55 & 0.51 & 0.22 \\
& Zero-shot w/ Hint & U-Shape & 0.55 & 0.47 & 0.35 & 0.51 \\
& Few-shot w/ CoT & Positive & 0.48 & 0.48 & 0.53 & 0.65 \\
\bottomrule
\end{tabular}

\vspace{2em}

\begin{tabular}{ccc|cccccc}
\toprule
Model & Prompting & Shape & ada & babbage & curie & davinci & davinci-v2 & davinci-v3 \\
\midrule
\multirow{3}{*}{GPT-3 Text Series} & Zero-shot & U-Shape & 0.61 & 0.53 & 0.48 & 0.31 & 0.56 & 0.71 \\
& Few-shot & U-Shape & 0.52 & 0.44 & 0.45 & 0.09 & 0.79 & 0.80 \\
& Zero-shot w/ Hint & U-Shape & 0.54 & 0.51 & 0.48 & 0.49 & 0.88 & 0.86 \\
& Few-shot w/ CoT & Positive & 0.34 & 0.45 & 0.47 & 0.89 & 0.98 & 0.98 \\
\bottomrule
\end{tabular}

\vspace{2em}

\begin{tabular}{ccc|cccc}
\toprule
Model & Prompting & Shape & small & medium & large & xlarge \\
\midrule
\multirow{3}{*}{Cohere} & Zero-shot & Inverse & 0.44 & 0.44 & 0.38 & 0.38 \\
& Few-shot & Inverse & 0.51 & 0.52 & 0.08 & 0.08 \\
& Zero-shot w/ Hint & U-Shape & 0.40 & 0.39 & 0.44 & 0.43 \\
& Few-shot w/ CoT & Positive & 0.47 & 0.47 & 0.75 & 0.75 \\
\bottomrule
\end{tabular}

\vspace{2em}

\begin{tabular}{ccc|ccc}
\toprule
Model & Prompting & Shape & large & grande & jumbo \\
\midrule
\multirow{3}{*}{Jurassic} & Zero-shot & Inverse & 0.59 & 0.49 & 0.49 \\
& Few-shot & U-Shape & 0.52 & 0.39 & 0.45 \\
& Zero-shot w/ Hint & Inverse & 0.58 & 0.48 & 0.44 \\
& Few-shot w/ CoT & Positive & 0.49 & 0.53 & 0.51 \\
\bottomrule
\end{tabular}

\caption{Table of performance corresponding to Figure~\ref{fig:scaling} and performance of few-shot prompting. }
\label{tab:performance}
\end{table*}

\begin{table*}[htbp]
\centering
\small

\begin{tabular}{c|ccc|cccccc}
\toprule
Task & Model & Prompting & Shape & ada & babbage & curie & davinci & d-v2 & d-v3 \\
\midrule
\multirow{2}{*}{Task 1} & GPT-3 & Zero-shot & Positive & 0.44 & 0.47 & 0.61 & 0.76 & - & - \\
& GPT-3 Text Series & Zero-shot & Positive & 0.41 & 0.47 & 0.51 & 0.88 & 0.94 & 0.95 \\
\midrule
\multirow{4}{*}{Task 2} & GPT-3 & Zero-shot & Sigmoid & 0.49 & 0.50 & 0.22 & 0.53 & - & - \\
& GPT-3 & Zero-shot w/ Hint & Sigmoid & 0.50 & 0.50 & 0.46 & 0.83 & - & - \\
& GPT-3 Text Series & Zero-shot & Sigmoid & 0.63 & 0.49 & 0.50 & 0.51 & 0.95 & 0.99 \\
& GPT-3 Text Series & Zero-shot w/ Hint & Sigmoid & 0.51 & 0.49 & 0.50 & 0.94 & 1.00 & 0.99 \\
\bottomrule
\end{tabular}

\caption{Table of performance corresponding to Figure~\ref{fig:task}. Task 1 is question answering and task 2 is negation understanding. d-v2 and d-v3 are abbreviations of davinci-v2 and davinci-v3, respectively.}
\label{tab:performance-2}
\end{table*}

\subsection{Data}

In Table~\ref{tab:data_generation_example}, we provide examples showing the data generation process of the NeQA dataset that was introduced in \S \ref{sec:dataset}. 

In Table~\ref{tab:data_1} and~\ref{tab:data_2}, we present a list of 100 data samples from NeQA that were utilized throughout the paper to examine scaling behaviours and task decomposition.

\begin{table*}
\small
\centering
\begin{tabular}{p{0.12\linewidth}p{0.4\linewidth}p{0.4\linewidth}}
\toprule
Dataset & Original Data & Transformed Data \\
\midrule
NegatedLAMA & \textit{\textbf{Original Question:}} Child wants? & \textit{\textbf{Question:}} Child does not want? \textit{(take \textbf{Negated Question})} \\
\scriptsize{(including ConceptNet, GoogleRE, SQuAD, TREx)} & \textit{\textbf{Original Answer:}} love & \textit{\textbf{Choices:}} [love, marriage] \textit{(take \textbf{Original Answer} and misprime in \textbf{Misprimed Question})} \\
& \textit{\textbf{Negated Question:}} Child does not want? & \textit{\textbf{Answer:}} marriage \textit{(take misprime in \textbf{Misprimed Question})}\\
& \textit{\textbf{Misprimed Question:}} Marriage? Child wants? \\
\midrule
OBQA & \textit{\textbf{Original Question:}} Pushing on a pedal is an example of? & \textit{\textbf{Question:}} Pushing on a pedal isn't an example of? \textit{(apply rule-based transformation on \textbf{Original Question})}  \\
& \textit{\textbf{Original Choices:}} [patching, force, practice, speed] & \textit{\textbf{Choices:}} [force, speed] \textit{(take \textbf{Original Answer} and sample one incorrect \textbf{Original Choices})}  \\
& \textit{\textbf{Original Answer:}} force & \textit{\textbf{Answer:}} speed \textit{(take sampled incorrect \textbf{Original Choices})} \\
\bottomrule
\end{tabular}
\caption{Data generation process of NeQA dataset. NeQA is constructed by transforming two existing QA datasets: NegatedLAMA and OBQA. All the fields of the original and transformed questions are shown.}
\label{tab:data_generation_example}
\end{table*}

\begin{table*}
\centering
\scriptsize
\begin{tabular}{p{0.62\linewidth}p{0.25\linewidth}l}
\toprule
Question & Choices & Answer \\
\midrule
Eden Roc Renaissance Hotel Miami Beach is not a? & [hotel, download] & 1 \\
The Betty Hutton Show was not originally aired on? & [youtube, cbs] & 0 \\
Hot 8 Brass Band does not play \_\_\_ music? & [jazz, friedrich] & 1 \\
Crying is not part of? & [life, love] & 1 \\
Melbourne International Film Festival is not located in? & [april, melbourne] & 0 \\
Mitsubishi G4M is not developed by? & [mitsubishi, vinci] & 1 \\
A man grows flowers in the winter using his greenhouse. The flowers blossom there, despite the cold, freezing weather and lack of sunshine, not because & [they are regularly chilled, they are carefully protected] & 0 \\
The main reason people want to use wind as an energy source is not because there is & [make cookies, an everlasting supply] & 0 \\
A light isn't refracted when & [it shines at a mirror, it shines at a gem] & 0 \\
John Selden did not die in? & [manila, london] & 0 \\
adenine does not consist of? & [chocolate, nitrogen] & 0 \\
Kolkata Book Fair is not located in? & [chinatown, kolkata] & 0 \\
Fall is not? & [coming, fearing] & 1 \\
Maurice Joly is not a \_\_\_ by profession? & [barrister, lawyer] & 0 \\
Sojitra is not a? & [village, town] & 1 \\
Avoiding the use of fossil fuel is not an example of & [Ecosystem Vulnerability, Conservationism] & 0 \\
The native language of Jean-Paul Sartre is not? & [french, greek] & 1 \\
Birds cannot? & [fly, ulster] & 1 \\
Worms are excellent decomposers not because & [worms enjoy eating dirt and decay, they enjoy eating carcasses and expel waste back to earth] & 0 \\
The capital of Azerbaijan is not? & [baku, kidnapped] & 1 \\
An electric car uses less gasoline than a regular car not because & [gas is unable to be used, power is prioritized over gas] & 0 \\
chicken nugget does not consist of? & [beef, chicken] & 0 \\
The native language of Eugene Kaspersky is not? & [russian, korean] & 1 \\
Digital Audio Tape is not owned by? & [sony, parsons] & 1 \\
Lizards do not have a? & [tail, song] & 1 \\
When the moon, sun, and earth line up, tides aren't at their & [Happy, Max] & 0 \\
Jean Nicolas Pierre Hachette did not die in? & [navarro, paris] & 0 \\
Rottendorf is not located in? & [germany, croatia] & 1 \\
David Sainsbury, Baron Sainsbury of Turville never worked in? & [london, director] & 1 \\
ground beef is not a subclass of? & [beef, pigs] & 1 \\
President of Ireland is not a legal term in? & [irish, ireland] & 0 \\
The official language of Ruovesi is not? & [wrestled, finnish] & 0 \\
A creature can hide in plain sight from predators not because & [it misdirects, it burns] & 1 \\
Information is not a? & [beauty, commodity] & 0 \\
The native language of Jules Roy is not? & [french, polish] & 1 \\
disruptive coloration is not a subclass of? & [insects, camouflage] & 0 \\
Organisms cannot? & [teatro, die] & 0 \\
Abdel Sattar Sabry does not play in \_\_\_ position? & [wing, midfielder] & 0 \\
Some animals get caught easily by other animals not because they are slow while the predators are & [manipulative, speedy] & 0 \\
Ano Liosia Olympic Hall is not located in? & [strikeouts, athens] & 0 \\
Esko Aho does not work for? & [nintendo, nokia] & 0 \\
Riga and \_\_\_ are not twin cities? & [tallinn, augsburg] & 1 \\
OKK Beograd is not located in? & [homosexual, belgrade] & 0 \\
Canjuers is not located in? & [bordeaux, france] & 0 \\
Child does not want? & [lab, love] & 0 \\
Tungsten is a metal which is found in large amounts in filaments, which are used in incandescent light bulbs. These light bulbs work not because & [the filament glows warmly, light bulbs get very warm] & 1 \\
Apple Thunderbolt Display is not developed by? & [intel, apple] & 0 \\
Funkadelic does not play \_\_\_ music? & [strips, funk] & 0 \\
Morice River is not a? & [muscle, river] & 0 \\
one inherited characteristic of fur isn't it's & [hue, weight] & 1 \\
\bottomrule
\end{tabular}
\caption{(1/2) Data samples from NeQA we used for scaling trends and task decomposition analyses. Note: labels are balanced.}
\label{tab:data_1}
\end{table*}

\begin{table*}
\centering
\scriptsize
\begin{tabular}{p{0.62\linewidth}p{0.25\linewidth}l}
\toprule
Question & Choices & Answer \\
\midrule
Salvadoran Football Federation is not a member of? & [fifa, uefa] & 1 \\
Biblioteka Dlya Chteniya was not written in? & [russian, helpful] & 1 \\
Pete Lesperance does not play? & [guitar, trombone] & 1 \\
Ohio House of Representatives is not a legal term in? & [delaware, ohio] & 0 \\
Jonathan Haidt does not work in the field of? & [music, psychology] & 0 \\
Internet Archive was not written in? & [english, polish] & 1 \\
Teres Ridge is not located in? & [antarctica, europe] & 1 \\
Joseph Clay was not born in? & [ponce, philadelphia] & 0 \\
One of the reasons some species go extinct is not because predators & [go dancing with them, murder too many of them] & 0 \\
Cans cannot be? & [impressed, recycled] & 0 \\
Lev Naumov does not play? & [piano, harp] & 1 \\
Tom Cruise is not affiliated with the \_\_\_ religion? & [scientology, islam] & 1 \\
Anti-Oedipus does not work in the field of? & [philosophy, algebra] & 1 \\
Two wheeled conveyances are not held together by & [glue, helically ridged pins] & 0 \\
Louisiana Voodoo was not founded in? & [nagoya, louisiana] & 0 \\
Animator.ru was not created in? & [russia, argentina] & 1 \\
Dan Sealey does not play? & [guitar, pipe organ] & 1 \\
Football Association of Brunei Darussalam is not a member of? & [fifa, application] & 1 \\
The leading cause of soil and rock erosion is not & [NaCl, H2O] & 0 \\
Cyprus is not an? & [accessory, island] & 0 \\
Cape Town and \_\_\_ are not twin cities? & [johannesburg, jerusalem] & 1 \\
The Earth's closest heat source isn't & [our celestial fireball, gamma rays] & 1 \\
Cars are big polluters not because they release toxins from the gas into the air making it & [space, unhealthy to breathe] & 0 \\
Italy does not maintain diplomatic relations with? & [lebanon, insights] & 1 \\
Heribert of Cologne has not the position of? & [mayor, archbishop] & 0 \\
Parippally is not located in? & [propagation, india] & 0 \\
Which of these isn't less likely to cause pollution & [chopper, mountain bike] & 0 \\
National Film Board of Canada does not work in the field of? & [animation, art] & 1 \\
fermented milk product does not consist of? & [syntax, milk] & 0 \\
Quartz crystals aren't made up of & [hexagons, majic] & 1 \\
nickel silver does not consist of? & [shrine, copper] & 0 \\
Rats cannot? & [sadness, bite] & 0 \\
A landslide is not likely to occur on a slope as & [downward movement is facilitated, there is more debris] & 1 \\
Tire is not \_\_\_ wheels? & [rubber, culminated] & 1 \\
A person wanting to find a live bear in a forest will have difficulty not because bears & [are friendly, avoid humanity] & 0 \\
Corn and wheat aren't among the favorite foods of & [humans, fish] & 1 \\
Percussion instruments is not made of? & [wood, temple] & 1 \\
Chevrolet Trax is not produced by? & [specialist, chevrolet] & 0 \\
The original language of The Bitch is not? & [english, breton] & 1 \\
Bill Gates does not work for? & [microsoft, ibm] & 1 \\
Rajesh Khanna not used to communicate in? & [hindi, yiddish] & 1 \\
Edinburgh Airport is not named after? & [edinburgh, bourne] & 1 \\
William of Hirsau has not the position of? & [abbot, republic] & 1 \\
Austrian Football Association is not a member of? & [lecturer, fifa] & 0 \\
Daniele Franceschini was not born in? & [telegraph, rome] & 0 \\
The native language of Louis Barthou is not? & [french, mortar] & 1 \\
A thing's position is not altered when & [thing feels moved emotionally, the thing adjusts its location] & 0 \\
Name is not proper? & [noun, address] & 1 \\
Which isn't recyclable & [Styrofoam food container, soda can] & 0 \\
The Jazztet does not play \_\_\_ music? & [jazz, classical] & 1 \\
\bottomrule
\end{tabular}
\caption{(2/2) Data samples from NeQA we used for scaling trends and task decomposition analyses. Note: labels are balanced.}
\label{tab:data_2}
\end{table*}

\subsection{Prompts}

The specific prompts utilized for various prompting methods and tasks are outlined in Table~\ref{tab:prompt}.

\begin{table*}[htbp]
\centering
\scriptsize
\begin{tabular}{l|p{0.7\linewidth}}
\toprule
Zero-Shot & The following are multiple choice questions (with answers) about common sense. \\
\\
& Question: Child does not want? \\
& A. love \\
& B. marriage \\
& Answer: \\

\midrule
Zero-Shot w/ Hint & The following are multiple choice questions (with answers) about common sense. Note that if there is a negation in the question, we should choose the wrong answer to the original question. \\
\\
& Question: Child does not want? \\
& A. love \\
& B. marriage \\
& Answer: \\

\midrule

Few-Shot w/ CoT & The following are multiple choice questions (with answers) about common sense. \\
\\
& Question: If a cat has a body temp that is below average, it isn't in \\
& A. danger \\
& B. safe ranges \\
& Answer: Let's think step-by-step. \\
& First, let's answer non-negated question: "If a cat has a body temp that is below average, it is in?" \\
& For this non-negated question, we have A. \\
& Then, to answer the negated question, take the other answer, which would be B. \\
& So the answer is B. \\
\\
& Question: As the barometer reading goes lower there is not a greater chance of \\
& A. sunshine \\
& B. getting wet \\
& Answer: Let's think step-by-step. \\
& First, let's answer non-negated question: "As the barometer reading goes lower there is a greater chance of?" \\
& For this non-negated question, we have B. \\
& Then, to answer the negated question, take the other answer, which would be A. \\
& So the answer is A. \\
\\
& Question: Coral is a type of living organism which cannot be identified in \\
& A. saltwater locations that are open \\
& B. any where with liquid \\
& Answer: Let's think step-by-step. \\
& First, let's answer non-negated question: "Coral is a type of living organism which can be identified in?" \\
& For this non-negated question, we have A. \\
& Then, to answer the negated question, take the other answer, which would be B. \\
& So the answer is B. \\
\\
& Question: Child does not want? \\
& A. love \\
& B. marriage \\
& Answer: \\

\midrule
\midrule
Task 1 & The following are multiple choice questions (with answers) about common sense. \\
\\
& Question: Child wants? \\
& A. love \\
& B. marriage \\
& Answer: \\
\midrule
\midrule
Task 2 (Weaker Prompt) & Sentence 1: "Child wants love." \\
& Sentence 2: "Child does not want love." \\
& Question: The above two sentences are? \\
& A. the same \\
& B. different \\
& Answer: \\
\midrule
Task 2 (Stronger Prompt) & Negated sentences are different from original sentences. \\
\\
& Sentence 1: "Child wants love." \\
& Sentence 2: "Child does not want love." \\
& Question: The above two sentences are? \\
& A. the same \\
& B. different \\
& Answer: \\
\bottomrule
\end{tabular}
\caption{Specific prompts for various prompting methods and tasks. Note: for Task 2 prompts, we randomly swap labels ``the same'' and ``different'' to balance the distribution. }
\label{tab:prompt}
\end{table*}

\subsection{Models}
\label{sec:models}

In Table~\ref{tab:model}, we present a list of all the models used in this work, including 4 model families and 17 models. Model details are from~\citet{liang2022holistic}.

\begin{table*}[htbp]
\centering
\small
\begin{tabular}{llp{0.7\linewidth}}
\toprule
Family & Model & Details \\
\midrule
\multirow{4}{*}{GPT-3} &	ada &	Original GPT-3 (350M parameters) autoregressive language model. \\
 &	babbage &	Original GPT-3 (1.3B parameters) autoregressive language model. \\
 &	curie &	Original GPT-3 (6.7B parameters) autoregressive language model. \\
 &	davinci &	Original GPT-3 (175B parameters) autoregressive language model. \\
\midrule
\multirow{10}{*}{GPT-3 Text Series} &	ada &	text-ada-001 model that involves supervised fine-tuning on human-written demonstrations. \\
 &	babbage &	text-babbage-001 model that involves supervised fine-tuning on human-written demonstrations. \\
 &	curie &	text-curie-001 model that involves supervised fine-tuning on human-written demonstrations. \\
 &	davinci-v1 &	text-davinci-001 model that involves supervised fine-tuning on human-written demonstrations. \\
 &	davinci-v2 &	text-davinci-002 model that involves supervised fine-tuning on human-written demonstrations. Derived from code-davinci-002. \\
 &	davinci-v3 &	text-davinci-003 model that involves reinforcement learning (PPO) with reward models. Derived from text-davinci-002. \\
\midrule
\multirow{4}{*}{Cohere} &	small &	Cohere small v20220720 (410M parameters). \\
 &	medium &	Cohere medium v20220720 (6.1B parameters). \\
 &	large &	Cohere large v20220720 (13.1B parameters). \\
 &	xlarge &	Cohere xlarge v20220609 (52.4B parameters). \\
\midrule
\multirow{3}{*}{Jurassic} &	large &	Jurassic-1 Large (7.5B parameters). \\
 &	grande &	Jurassic-1 Grande (17B parameters) with a few tweaks to the training process. \\
 &	jumbo &	Jurassic-1 Jumbo (178B parameters). \\
\bottomrule
\end{tabular}
\caption{List of models used in this work, including 4 model families and 17 models. Note that the publicly available GPT-3 Text Series model APIs used in this paper differ from those described in the original InstructGPT paper~\cite{ouyang2022training}, and OpenAI does not provide information on the training procedure and appropriate model sizes~\cite{liang2022holistic}. 
}
\label{tab:model}
\end{table*}
\section{Additional Analyses}

\subsection{Few-Shot Prompting} 

Few-shot in-context learning has been demonstrated to be an effective method for adapting pre-trained language models to specific tasks. We experimented with few-shot prompting (not few-shot chain-of-thought prompting) but didn't include the results in the main paper because the scaling shapes were often the same as zero-shot prompting across 3 out of 4 studied model families (inverse for GPT-3 and Cohere, U-shape for GPT-3 Text-Series; only Jurassic changes from inverse to U-shape). We provide the few-shot prompting results in Table~\ref{tab:performance}.

Several recent works can explain why few-shot prompting doesn't alter the scaling curve shape. For example, \citet{min2022rethinking} and \citet{xie2021explanation} show that in-context learning can be viewed as a Bayesian inference process, with the model learning more about input-output format than input-output mapping. When providing demonstrations of negated question-answer pairs, the model fails to learn the mapping between them and predicts the same answer as without demonstrations.

\subsection{Prompt Variations}

Due to the sensitivity of language model performance to prompts~\cite{jiang-etal-2020-know,ruis2022large,wang2022self} (also discussed in limitations), we experimented with various prompts and found:

\begin{enumerate}
\item Minor changes like word substitution or paraphrasing result in similar scaling shapes;
\item Major prompt changes can alter curve shape, e.g., adding `For example, ``isn't'', ``is not'', ``not because '', ``do not'' are strong signs of negation' to zero-shot w/ hint prompting changes GPT-3 from inverse to a weak U-shape. This can be seen as increasing CoT strength by providing more hints/rationales;
\item Varying CoT information levels affects the shape. Intermediate-level information in CoT prompts shows a scaling shape between U-shape (zero-shot w/ hint; weakest CoT version) and strong positive (few-shot CoT; strongest CoT version).
\end{enumerate}

\subsection{Dataset Validity}

NeQA is curated by applying rule-based transformations on existing QA datasets. To ensure the dataset quality, we carefully design and verify the transformation rules through manual inspection of the transformed examples. We found that the transformation rules generally work well and only removed a few questions due to grammatical errors after adding negation.

Furthermore, as part of the submission for the inverse scaling prize~\cite{mckenzie2022inverse}, the organizers have done crowdsourcing experiments to demonstrate the validity of our dataset. Specifically, they validated labels by crowdsourcing 50 random examples from NeQA, and found the average agreement between workers and gold labels is 100\% with no confusing questions.

\subsection{Subset Selection}

\begin{figure*}[!b]
\begin{centering}
    \begin{minipage}[htbp]{0.32\linewidth}
        \centering
        \includegraphics[width=\linewidth]{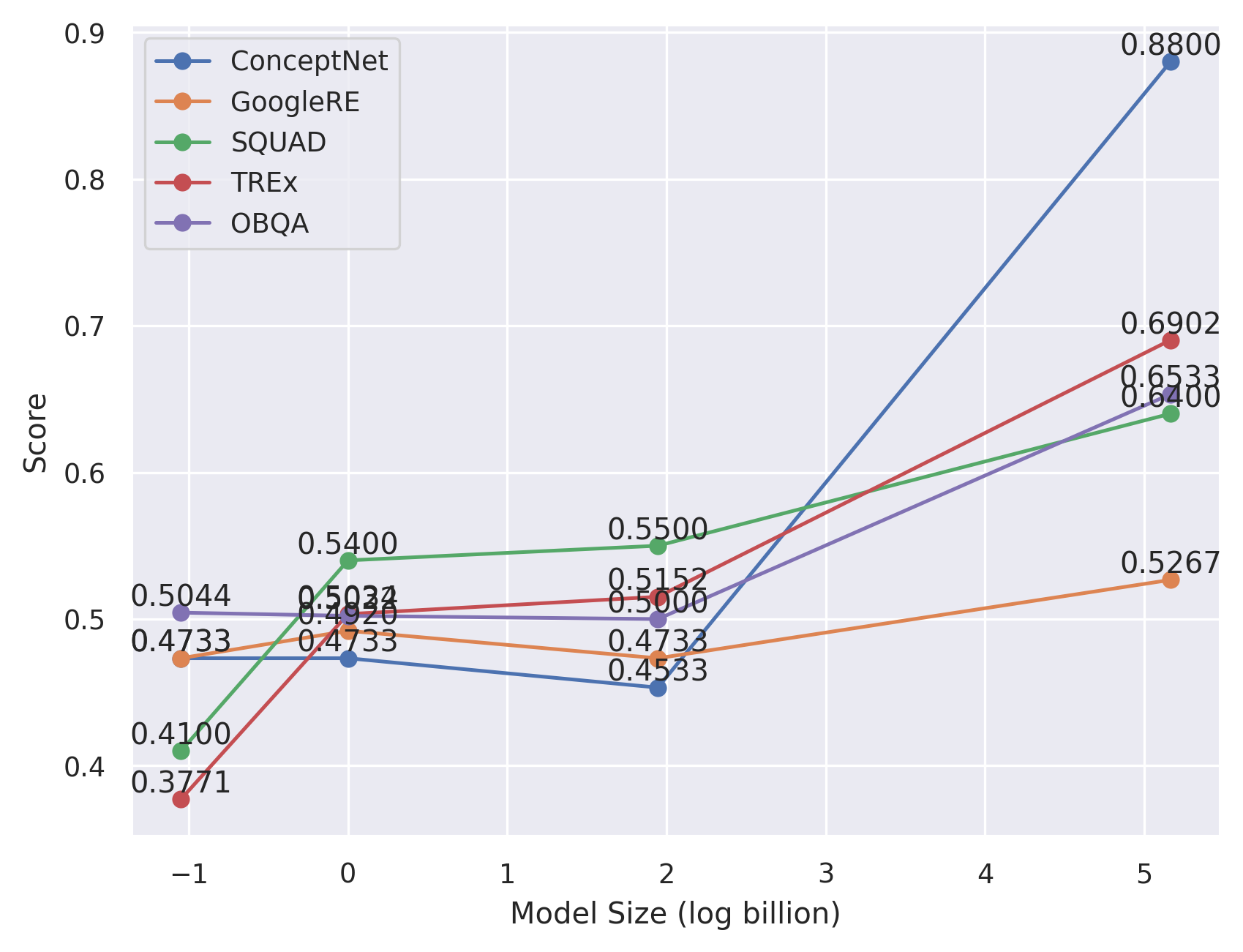}
    \end{minipage}
    \begin{minipage}[htbp]{0.32\linewidth}
        \centering
        \includegraphics[width=\linewidth]{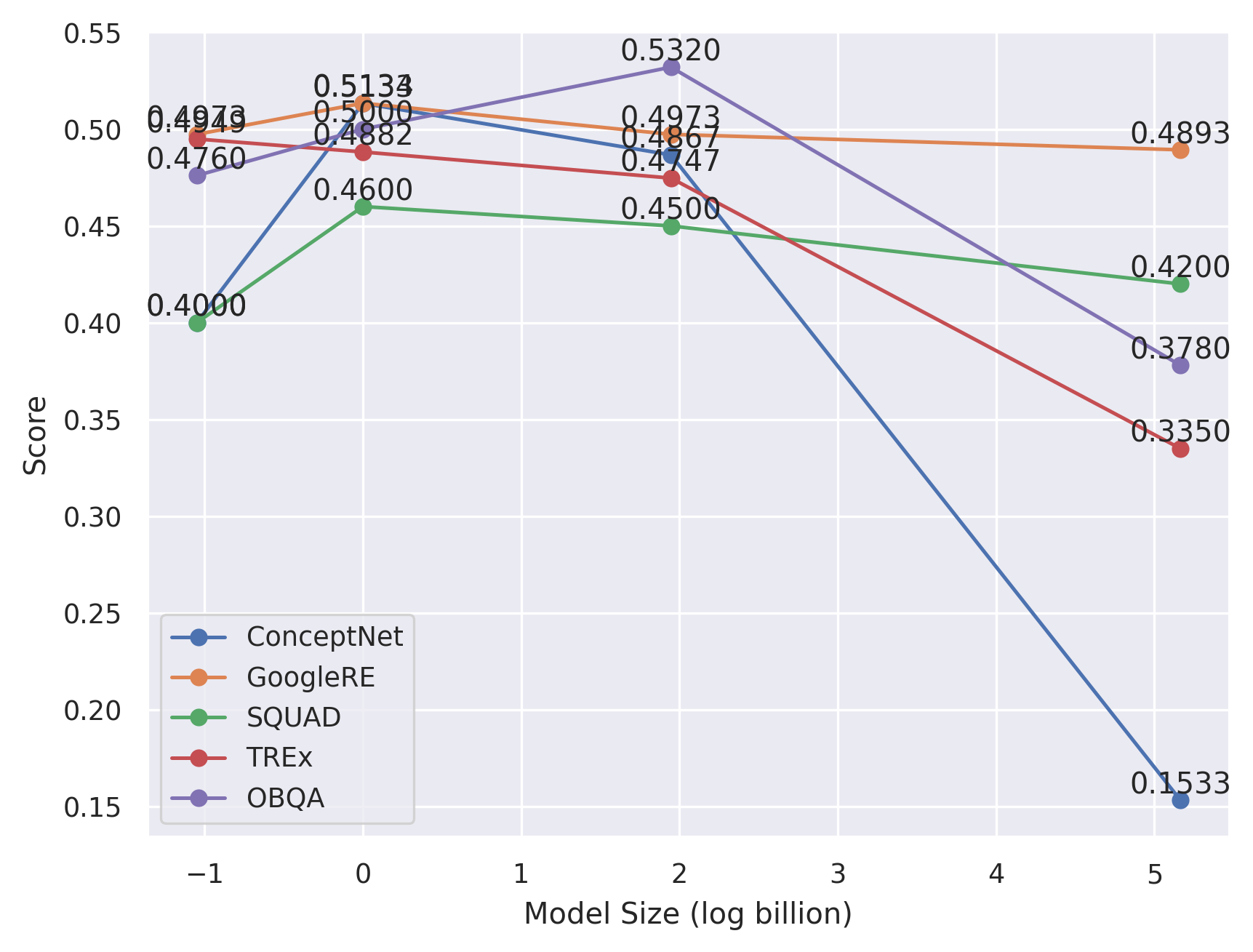}
    \end{minipage}    
    \caption{Scaling trend of GPT-3 on five subsets in NeQA using zero-shot prompting. Left: positive scaling (i.e., Task 1) on original questions. Right: inverse scaling on negated questions. Since GPT-3 does not exhibit strong positive scaling and inverse scaling on the original and negated GoogleRE and SQUAD datasets, these subsets were not included in the analysis.}
    \label{fig:inverse_scaling}
\end{centering}
\end{figure*}

The NeQA dataset is composed of five subsets: ConceptNet, GoogleRE, SQUAD, TREx, and OBQA. For the purpose of this analysis, we only include ConceptNet, TREx, and OBQA. Our goal is to examine the scaling trends, so we aim for steeper scaling. However, GPT-3 does not exhibit strong positive scaling and inverse scaling on the original and negated GoogleRE and SQUAD datasets (Figure~\ref{fig:inverse_scaling}), so these subsets were not included in the analysis. 

Furthermore, these scaling trends of NeQA subsets provide additional verification of our task decomposition analysis. When language models fail to understand negation (Task 2), a stronger positive scaling on the original dataset (Task 1) causes a stronger inverse scaling on the negated data (Composed Task).

\subsection{Negation Category} 

In Figure~\ref{fig:negation_analysis} (left), we find that negating by adding ``un-/in-'' prefix to a word or negating modal verbs (e.g., ``can'' to ``cannot'') does not show clear inverse scaling in zero-shot prompting. We suspect that the difference is because these negation categories replace a word instead of adding an additional word ``not''. We leave the further analysis to future work.

\subsection{Wrong Choice}

This experiment aimed to understand whether more confusing choices will change the scaling (Figure~\ref{fig:negation_analysis} (middle)). For example, given the question ``Apple is not made by'', the wrong choice can be ``Microsoft'' (high-ranked, more confusing), or ``air'' (low-ranked, less confusing), or a random word ``China''. We find that the wrong choice has little impact on the scaling trends.

\subsection{Mispriming}

\begin{figure*}
\begin{centering}
    \begin{minipage}[htbp]{0.32\linewidth}
        \centering
        \includegraphics[width=\linewidth]{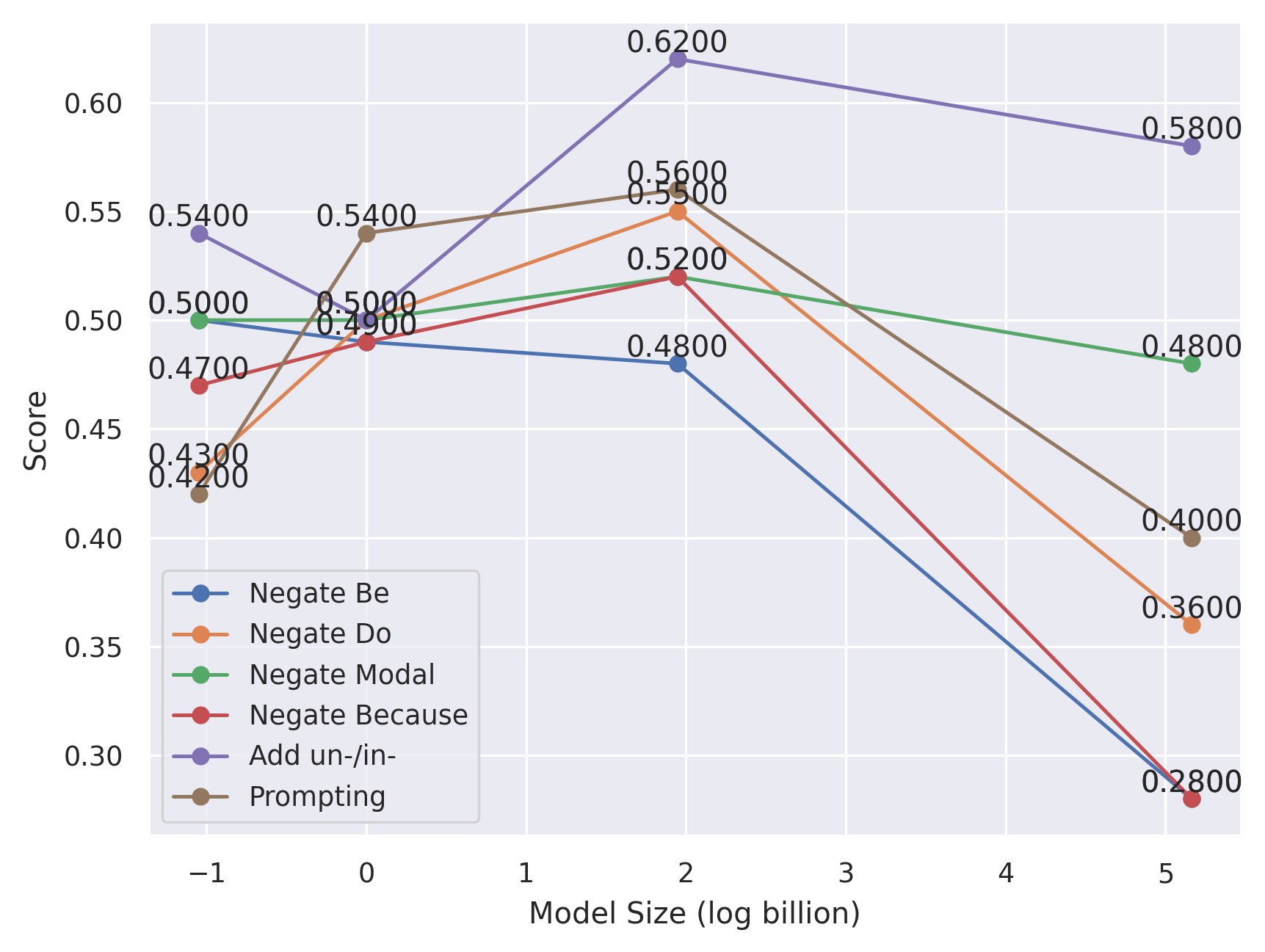}
    \end{minipage}
    \begin{minipage}[htbp]{0.32\linewidth}
        \centering
        \includegraphics[width=\linewidth]{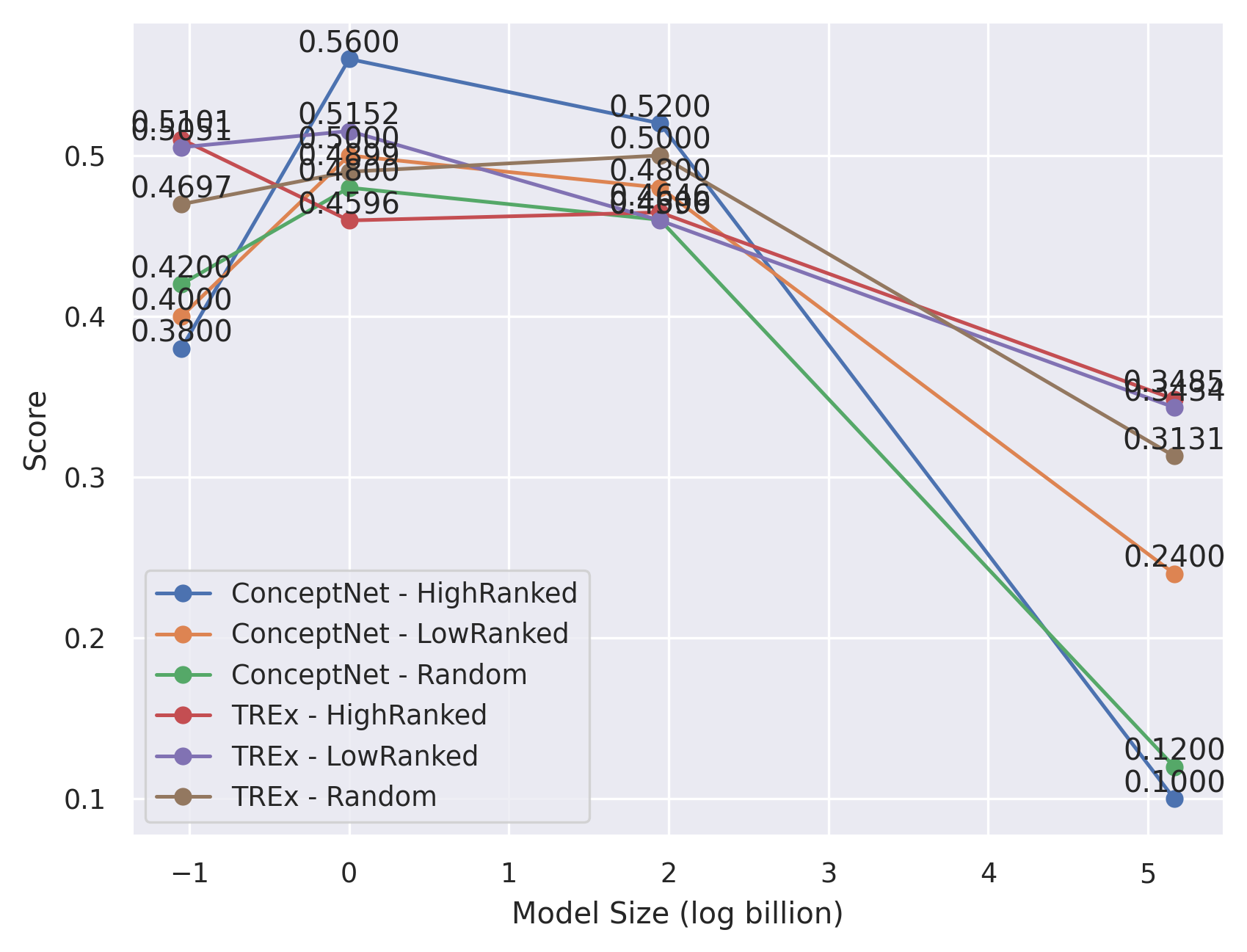}
    \end{minipage}    
    \begin{minipage}[htbp]{0.32\linewidth}
        \centering
        \includegraphics[width=\linewidth]{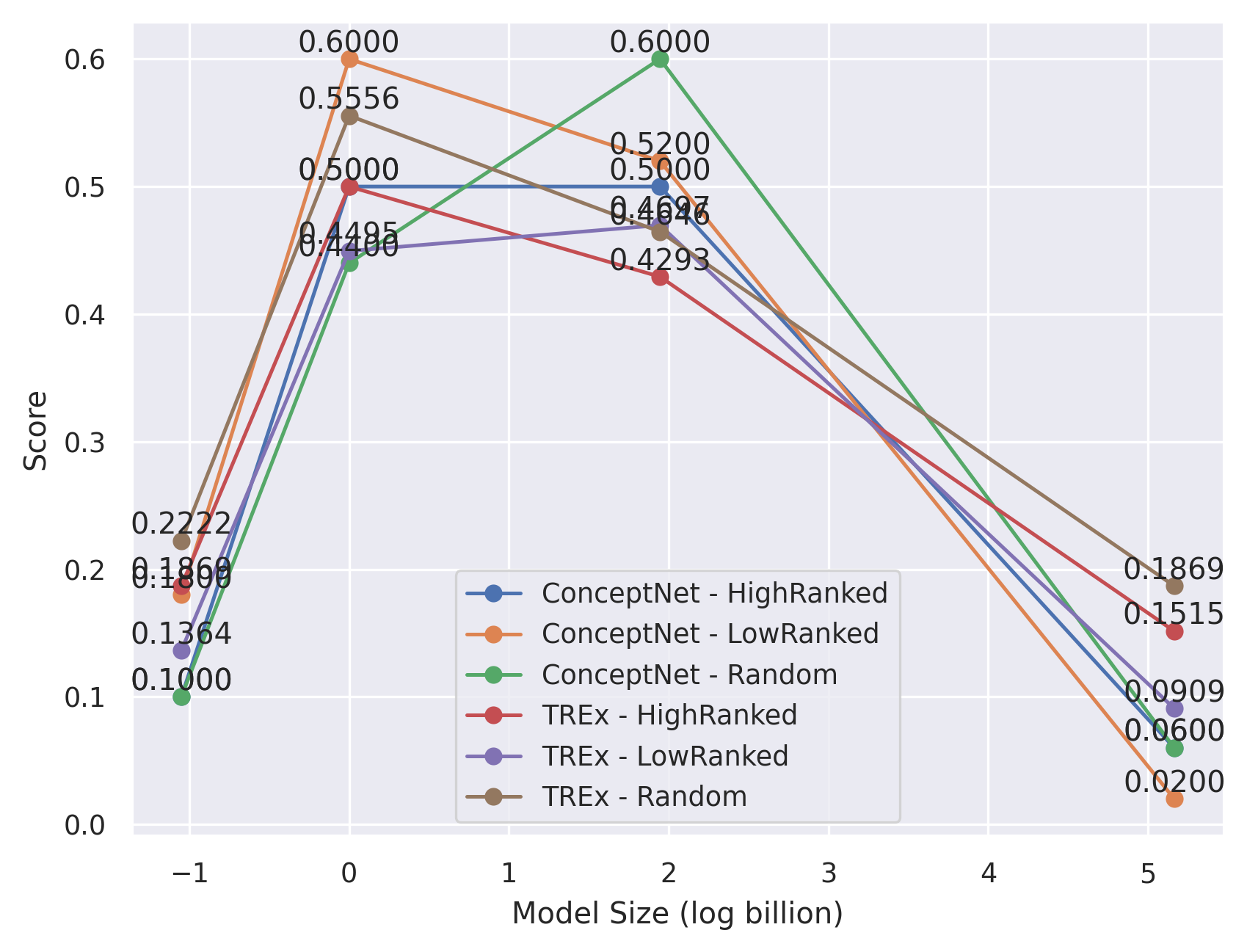}
    \end{minipage}
    \caption{The impact of negation category, wrong choice, and mispriming on scaling trend. These experiments are done using GPT-3 and zero-shot prompting.}
    \label{fig:negation_analysis}
\end{centering}
\end{figure*}

Following~\citet{kassner-schutze-2020-negated}, we put the wrong choice (i.e., the correct choice before negation) before the question (e.g., change ``iPhone is not made by'' to ``Apple? iPhone is not made by''). Mispriming makes inverse scaling stronger on negated questions in zero-shot prompting setting (Figure~\ref{fig:negation_analysis} (right)). Interestingly, we also note a phase change happens in small-size models. While this is a very interesting finding, mispriming might not be frequent in real-world applications of language models, so we are not including this in the NeQA dataset.

\end{document}